%% file: acl_latex.tex
\pdfoutput=1

\documentclass[11pt]{article}

\usepackage[preprint]{acl}

\usepackage{times}
\usepackage{latexsym}

\usepackage[T1]{fontenc}

\usepackage[utf8]{inputenc}

\usepackage{microtype}

\usepackage{inconsolata}

\usepackage{graphicx}

\usepackage{hyperref}       
\usepackage{url}            
\usepackage{booktabs}       
\usepackage{amsfonts}       
\usepackage{nicefrac}       
\usepackage{microtype}      
\usepackage{lipsum}
\usepackage{fancyhdr}       
\usepackage{graphicx}       
\usepackage{float}
\usepackage{multirow}
\usepackage[utf8]{inputenc}
\graphicspath{{media/}}     
\usepackage{siunitx} 
\usepackage{gensymb} 
\usepackage[most]{tcolorbox}

\usepackage{markdown}
\usepackage{tabularx}
\usepackage{xcolor}
\usepackage{hyperref}
\usepackage{multirow}
\usepackage{listings}
\usepackage{longtable}
\usepackage{bbding}
\usepackage{enumitem}
\usepackage{amssymb}
\usepackage{wrapfig}
\usepackage{array}
\hypersetup{
    colorlinks=true, 
    linkcolor=blue, 
    filecolor=magenta, 
    urlcolor=cyan, 
    citecolor=green 
}

\newtcolorbox{prompt}[1]{
    enhanced,
    colback=gray!20,
    colframe=black,
    boxrule=0.3pt,
    arc=3mm,
    left=2pt,
    right=2pt,
    boxsep=3pt,
    fonttitle=\small\bfseries,
    title=#1,
    fontupper=\scriptsize
}

\newcommand{\ie}{\emph{i.e., }}
\newcommand{\eg}{\emph{e.g., }}

\newcommand{\cf}{\emph{cf. }}

\definecolor{lightgray}{gray}{0.9}
\DeclareUnicodeCharacter{FF08}{\textlparen} 
\DeclareUnicodeCharacter{FF09}{\textrparen} 

\title{SciAssess: Benchmarking LLM Proficiency in Scientific Literature Analysis}

\author{%
\parbox{\linewidth}{\centering%
\textbf{Hengxing Cai$^{1}$\thanks{Equal Contribution}}, \textbf{Xiaochen Cai$^{1}$\footnotemark[1]}, \textbf{Junhan Chang$^{1}$\footnotemark[1]}, \textbf{Sihang Li$^1$\footnotemark[1]}, \textbf{Lin Yao$^1$}, \\[0.5ex] \textbf{Changxin Wang$^1$},  \textbf{Zhifeng Gao$^1$},  \textbf{Hongshuai Wang$^1$}, \textbf{Yongge Li$^1$}, \textbf{Mujie Lin$^1$}, \\[0.5ex]
\textbf{Shuwen Yang$^1$}, \textbf{Jiankun Wang$^1$}, \textbf{Mingjun Xu$^1$}, \textbf{Jin Huang$^1$}, \textbf{Xi Fang$^1$}, \textbf{Jiaxi Zhuang$^1$}, \\[0.5ex]
\textbf{Yuqi Yin$^1$}, \textbf{Yaqi Li$^1$}, \textbf{Changhong Chen$^1$}, \textbf{Zheng Cheng$^2$}, \textbf{Zifeng Zhao$^2$}, \\[0.5ex]
\textbf{Linfeng Zhang$^{1,2}$} and \textbf{Guolin Ke$^{1}$} \\[2ex]
$^1$DP Technology \quad $^2$AI for Science Institute, Beijing\\[2ex]
}%
}

\begin{document}
\maketitle
\input{chapters/0_abstract}


\input{chapters/1_introduction}
\input{chapters/2_benchmark}
\input{chapters/3_experiment}
\input{chapters/4_related_work}
\input{chapters/5_conclusion}

\newpage

\bibliography{custom}

\input{chapters/6_appendix}

\end{document}

%% file: chapters/0_abstract.tex
\begin{abstract}

Recent breakthroughs in Large Language Models (LLMs) have revolutionized scientific literature analysis. 
However, existing benchmarks fail to adequately evaluate the proficiency of LLMs in this domain, particularly in scenarios requiring higher-level abilities beyond mere memorization and the handling of multimodal data.
In response to this gap, we introduce SciAssess, a benchmark specifically designed for the comprehensive evaluation of LLMs in scientific literature analysis. 
It aims to thoroughly assess the efficacy of LLMs by evaluating their capabilities in Memorization (L1), Comprehension (L2), and Analysis \& Reasoning (L3). 
It encompasses a variety of tasks drawn from diverse scientific fields, including biology, chemistry, material, and medicine.
To ensure the reliability of SciAssess, rigorous quality control measures have been implemented, ensuring accuracy, anonymization, and compliance with copyright standards. 
SciAssess evaluates 11 LLMs, highlighting their strengths and areas for improvement. 
We hope this evaluation supports the ongoing development of LLM applications in scientific literature analysis.
SciAssess and its resources are available at \url{https://github.com/sci-assess/SciAssess}.

\end{abstract}

%% file: chapters/1_introduction.tex
\section{Introduction}
\begin{figure}
  \centering
  \includegraphics[width=\columnwidth]{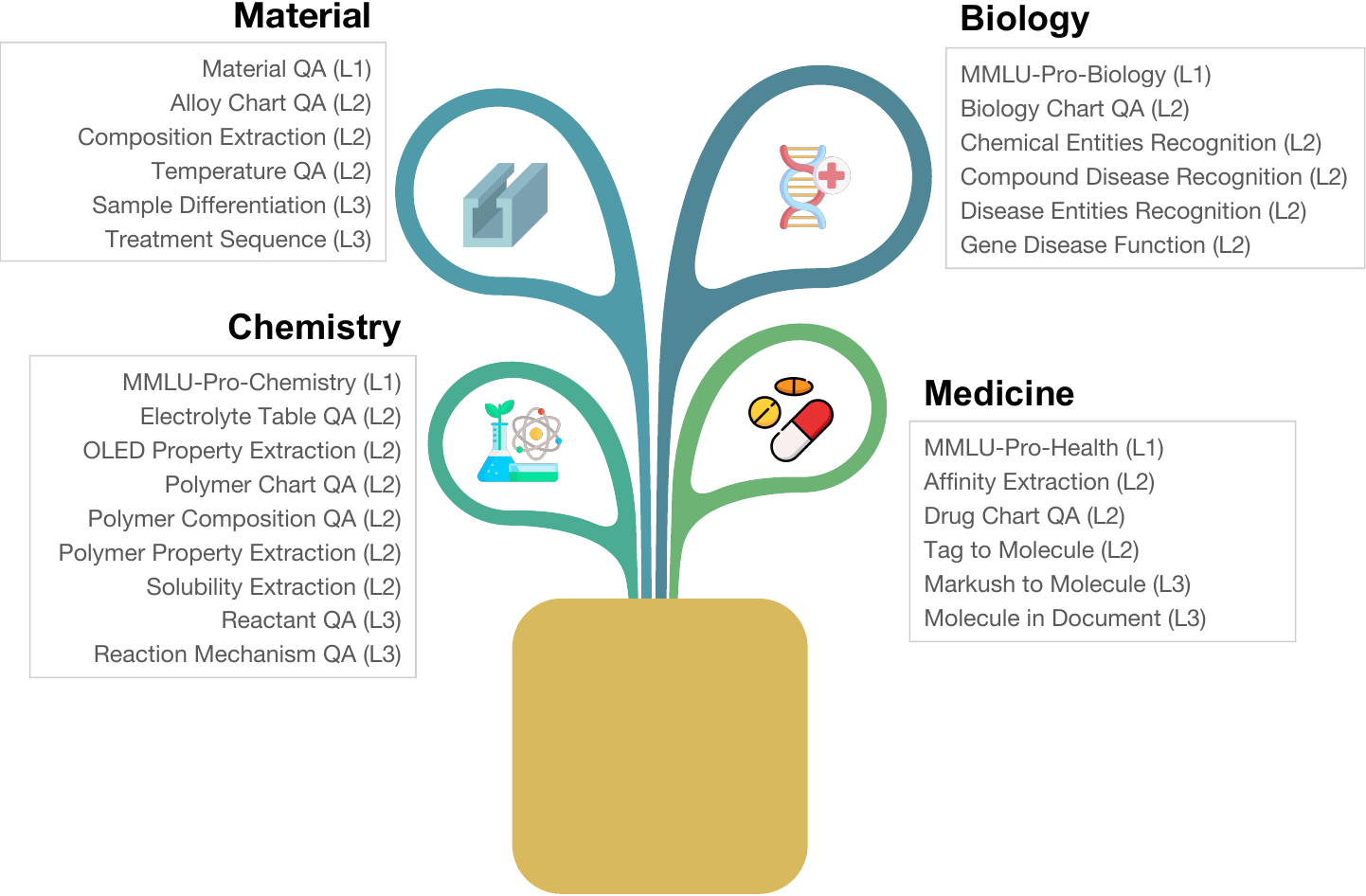} 
  \vspace{-8pt}
  \caption{Overview of SciAssess. It spans over 4 sub-domains and encompasses 27 tasks.}
  \label{fig:overview}
\end{figure}
Recent advances in Large Language Models (LLMs), such as GPT \cite{gpt4,gpt3}, Gemini \cite{gemini}, and Llama \cite{llama}, have attracted considerable attention due to their profound capabilities in natural language understanding and generation \cite{gpt-4-eval}. 
Evaluating these models is crucial for exploring their capability boundaries and limitations, thereby driving technological advancements.
In response, a variety of benchmarks tailored for LLMs have been proposed for extensive evaluation, covering a wide range of skills \cite{bm-2,bm-3} and diverse tasks \cite{bm-4, bm-5}.

\begin{figure*}[t]
    \centering
    \includegraphics[width=\textwidth]{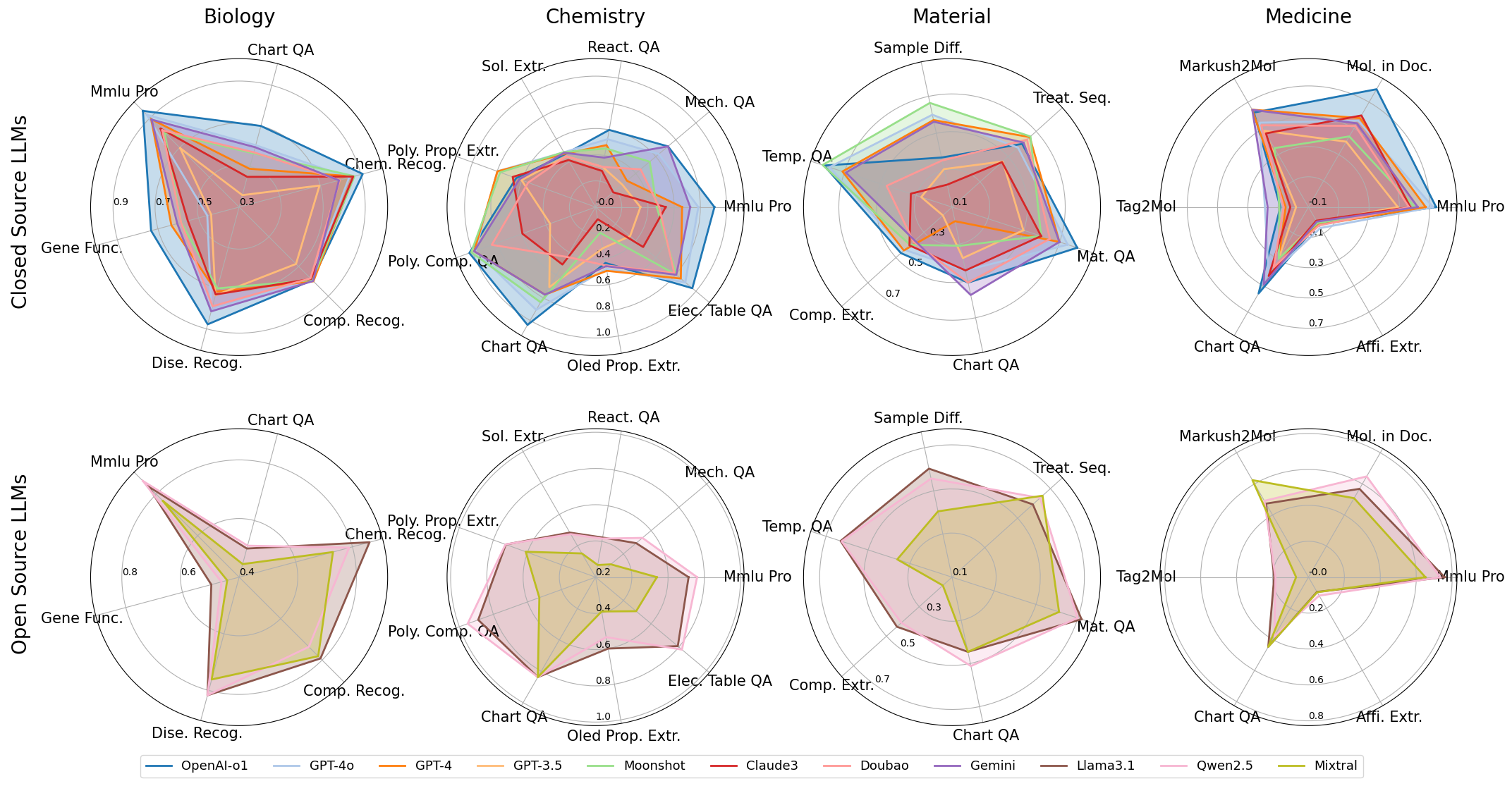}
    \vspace{-8pt}
    \caption{Performance overview of leading open and closed source LLMs on SciAssess. 
    Each column represents a scientific domain. 
    LLMs are evaluated on multiple tasks within each domain, with task details provided in Table \ref{tab:benchmark-stat}. 
    For closed source LLMs (first row), GPT-4o and GPT-4 are the leading models. For open source LLMs (second row), Llama3 and Qwen2 emerge as the top models.}
    \label{fig:performance-overview}
\end{figure*}

Despite LLMs not yet fully replacing scientific researchers in generating creative discoveries, they have demonstrated substantial potential in enhancing researchers' efficiency in scientific literature analysis \cite{science-llm}. 
Specific applications such as automatic literature summarization and knowledge extraction have seen practical deployments, significantly boosting researchers' productivity and expanding the range of literature that can be effectively utilized \cite{scientific-literature-llm}.
Inspired by Bloom's Taxonomy \cite{taxonomy}, we systemize the requirements for scientific literature analysis assistants into three progressive levels: (1) \textbf{Memorization (L1)}: Establishing an extensive foundational knowledge base to accurately address common factual questions in various scientific domains; (2) \textbf{Comprehension (L2)}: Identifying, extracting, and understanding the core content of provided documents; and (3) \textbf{Analysis \& Reasoning (L3)}: Integrating extracted information with the existing knowledge base to perform logical reasoning and analysis.

Existing comprehensive LLMs benchmarks, such as MMLU-Pro \cite{mmlu,mmlu-pro}, include some tasks related to scientific data. 
However, these sub-tasks have two limitations: (1) they mostly focus on Memorization, neglecting higher-level abilities such as L2 and L3; (2) these tasks lack the evaluation of various multi-modal inputs (\eg charts, molecular structures, and tables), which are crucial in scientific literature. 

In light of these existing limitations, we introduce \textbf{SciAssess} (\cf Figure~\ref{fig:overview}) -- a benchmark specifically designed for scientific literature analysis. 
SciAssess not only broadens the evaluation scope to encompass a wider range of LLM capabilities but also extends beyond text to include the extraction and interpretation of multimodal contents.
Moreover, meticulous design is essential to creating evaluations that yield deep insights, ensure fairness across different LLMs. 
Consequently, SciAssess is founded on three critical considerations:

\textbf{Model Ability.} A benchmark must clearly delineate the desired capabilities and model the intrinsic relationships among them, facilitating a diagnostic understanding.
Thus, SciAssess evaluates across three progressive levels (\ie Memorization (L1), Comprehension (L2), and Analysis \& Reasoning (L3)) and five modalities (\ie texts, charts, chemical reactions, molecular structures, and tables).
Consequently, SciAssess yields nuanced and informative evaluation outcomes, pinpointing specific aspects where the examined models may fall short.
    
\textbf{Scope \& Task.} Benchmarks should encompass a broad array of scientific domains to ensure comprehensiveness. 
Within each domain, the selected tasks must authentically represent the typical challenges and scenarios characteristic of that field.
Consequently, SciAssess spans over 4 sub-domains (\ie biology, chemistry, material, and medicine) and encompasses 27 tasks, each carefully suggested or designed by domain experts according to their professional experience.

\textbf{Scale \& Quality Control.} The scale and quality of the benchmark must be impeccable to serve as a dependable basis for deriving accurate, actionable, and applicable insights. 
SciAssess contains 6,938 questions in total to ensure adequate scale.
Each question is transformed from existing datasets or manually curated by domain experts hired by us \footnote{All data collection, annotation, and quality control tasks were carried out by the authors (who are also employees of the company) as part of their job responsibilities, and therefore, they were not provided with any additional compensation.}.
Subsequently, expert cross-validation is performed to ensure correctness and reliability. 

Overall, SciAssess aims to reveal the performance of LLMs as a scientific literature analysis assistant, thereby identifying their strength and weaknesses. 
The insights gained from SciAssess could hopefully catalyze further enhancing the capabilities of LLMs in scientific literature analysis, ultimately contributing to the acceleration of scientific discovery and innovation.

%% file: chapters/2_benchmark.tex
\section{Benchmark Dataset}

\begin{table*}[t]
    \centering
    \scriptsize
    \begin{tabular}{>{\centering\arraybackslash}m{1.1cm} >{\centering\arraybackslash}p{3.2cm} >{\centering\arraybackslash}p{0.6cm} >{\centering\arraybackslash}p{1.2cm} >
    {\centering\arraybackslash}p{2.5cm} >{\centering\arraybackslash}p{2cm} >{\centering\arraybackslash}p{1.25cm}
    }
    \toprule
    Domain & Task & Ability & \# Questions & Question Type & Metric & Modality \\
    \noalign{\vskip -1pt}\midrule\noalign{\vskip -1pt}
    \multirow{6}{1.1cm}{\centering Biology} 
    & \textcolor{gray}{MMLU-Pro-Biology} & L1 & 717 & Multiple Choice & Accuracy & Text only \\
    & Biology Chart QA & L2 & 199 & Multiple Choice & Accuracy & Chart \\
    & \textcolor{gray}{Chemical Entities Recognition} & L2 & 500 & Text Extraction & F1-score & Text only \\
    & \textcolor{gray}{Compound Disease Recognition} & L2 & 775 & Text Extraction & F1-score & Text only \\
    & \textcolor{gray}{Disease Entities Recognition} & L2 & 500 & Text Extraction & F1-score & Text only \\
    & Gene Disease Function & L2 & 24 & Text Extraction & F1-score & Text only \\
    \noalign{\vskip -1pt}\midrule\noalign{\vskip -1pt}
    \multirow{9}{1.1cm}{\centering Chemistry} 
    & \textcolor{gray}{MMLU-Pro-Chemistry} & L1 & 1,132 & Multiple Choice & Accuracy & Text only \\
    & Electrolyte Table QA & L2 & 200 & Multiple Choice & Accuracy & Table \\
    & OLED Property Extraction & L2 & 13 & Table Extraction & Recall & Mol., Table \\
    & Polymer Chart QA & L2 & 15 & Multiple Choice & Accuracy & Chart \\
    & Polymer Composition QA & L2 & 209 & Multiple Choice & Accuracy & Text only \\
    & Polymer Property Extraction & L2 & 109 & Table Extraction & Recall & Table \\
    & Solubility Extraction & L2 & 100 & Table Extraction & Recall & Table \\
    & Reactant QA & L3 & 195 & Multiple Choice & Accuracy & Reaction \\
    & Reaction Mechanism QA & L3 & 22 & Multiple Choice & Accuracy & Reaction \\
    \noalign{\vskip -1pt}\midrule\noalign{\vskip -1pt}
    \multirow{6}{1.1cm}{\centering Material} 
    & \textcolor{gray}{Material QA} & L1 & 263 & Multiple Choice & Accuracy & Text only \\
    & Alloy Chart QA & L2 & 15 & Multiple Choice & Accuracy & Chart \\
    & Composition Extraction & L2 & 244 & Table Extraction & Recall & Table \\
    & Temperature QA & L2 & 207 & Multiple Choice & Accuracy & Text only \\
    & Sample Differentiation & L3 & 237 & Multiple Choice & Accuracy & Text only \\
    & Treatment Sequence & L3 & 202 & True/False & Accuracy & Text only \\
    \noalign{\vskip -1pt}\midrule\noalign{\vskip -1pt}
    \multirow{7}{1.1cm}{\centering Medicine} 
    & \textcolor{gray}{MMLU-Pro-Health} & L1 & 818 & Multiple Choice & Accuracy & Text only \\
    & Affinity Extraction & L2 & 40 & Table Extraction & Recall & Mol., Table \\
    & Drug Chart QA & L2 & 15 & Multiple Choice & Accuracy & Chart \\
    & Tag to Molecule & L2 & 50 & Mol. Generation & Mol. Similarity & Mol. \\
    & Markush to Molecule & L3 & 37 & Mol. Generation & Mol. Similarity & Mol. \\
    & Molecule in Document & L3 & 50 & True/False & Accuracy & Mol. \\
    \bottomrule
    \end{tabular}
    \caption{Statistics of the SciAssess. It comprises 6,888 questions across 27 tasks in five sub-domains. 
    Tasks are categorized into three ability levels: Memorization (L1), Comprehension (L2), and Analysis \& Reasoning (L3).
    Tasks that are \textcolor{gray}{gray} are transformed from existing datasets, while others are curated by domain experts hired by us.
    }
    \label{tab:benchmark-stat}
\end{table*}

We begin by outlining the ability assessment framework in Section \ref{sec:ability_assess}, which serves as the backbone of our evaluation framework. 
Moving forward, we provide detailed description of evaluation scopes and tasks in Section \ref{sec:scope_and_task}.
Lastly, we present the quality control measures implemented to ensure the integrity and reliability in Section \ref{sec:data_compliance}.

\subsection{Ability Assessment Framework}\label{sec:ability_assess}
Guided by the widely accepted cognitive learning processes outlined in Bloom's Taxonomy \cite{taxonomy}, we propose that the evaluation of LLMs in scientific literature analysis should be classified into three core levels:

\textbf{Memorization (L1)} refers to the model's extensive knowledge base, which allows it to accurately answer common factual questions in science autonomously.
\textbf{Comprehension (L2)} is the ability to precisely identify and extract key information and facts within a given text, and to comprehend them.
\textbf{Analysis \& Reasoning (L3)} demonstrate the model's advanced capability to amalgamate extracted information with its existing knowledge base for logical reasoning and analysis, leading to well-founded conclusions or predictions.

    
    

Inspecting existing LLM benchmarks in science field (See Section \ref{sec:related_work}) through three-level ability assessment framework, we find that they mostly focus on Memorization (L1) -- the foundational knowledge base for scientific facts -- while overlooking the higher-level abilities of Comprehension (L2) and Analysis \& Reasoning (L3).

Given the significant potential of leveraging LLMs as scientific literature analysis assistants to boost scientific discovery, we propose SciAssess as a more comprehensive benchmark, in terms of tasks, scopes, and modalities.


\subsection{Scope \& Task}\label{sec:scope_and_task}

After categorizing the ability of of LLMs into three levels, we proceed to introduce how we choose the tasks in SciAssess. 
First, we include four vertical domains: biology, chemistry, material, and medicine, as shown in Figure \ref{fig:overview}. 
This categorization ensures that SciAssess captures the unique challenges and requirements of each specific field.
Then, as mentioned above, Memorization (L1), being the extensive foundation for other higher-level abilities, should encompass as large a knowledge base as possible. 
Thus, SciAssess includes factual questions in 
MMLU-Pro~\cite{mmlu-pro} and MaScQA~\cite{mascqa}, covering fundamental knowledge in each field.
For the evaluation of Comprehension (L2) and Analysis \& Reasoning (L3), we identify realistic demands by consulting domain experts and curate corresponding tasks.
The reason is that solving tasks in these domains require finer-grained abilities, such as understanding tables and molecular structures.
For instance, crucial composition information in material science literature is often found in tables, whereas key information extraction in drug discovery necessitates the accurate recognition of molecular structures.

\begin{figure}[t]
  \centering
  \includegraphics[width=\columnwidth]{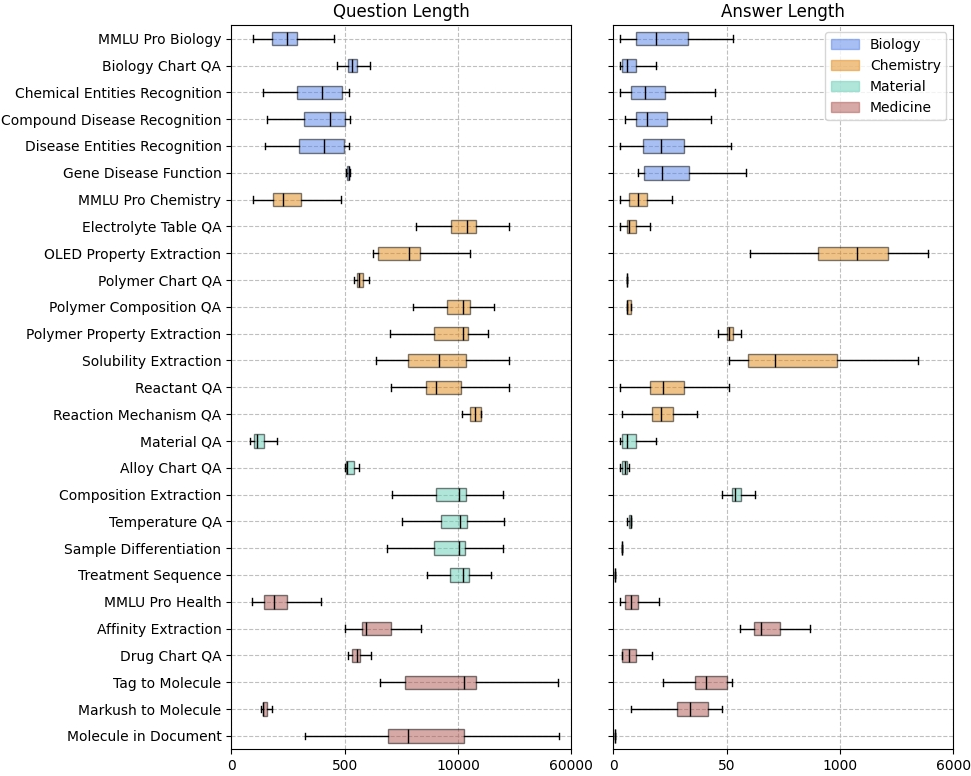} 
  \caption{Distribution of token length for questions and answers in each task.}
  \label{fig:length}
\end{figure}

SciAssess, as presented in Table \ref{tab:benchmark-stat}, comprises 6,888 questions across 27 tasks in five scientific domain, encompassing three ability levels: Memorization (L1), Comprehension (L2), and Analysis \& Reasoning (L3). 
Of these tasks, 7 out of 27 are transformed from existing public datasets (\textcolor{gray}{gray 
tasks} in Table~\ref{tab:benchmark-stat}) , and the other 21 tasks curated by us are based on contents from academic papers, specifically designed to assess the ability to analyze scientific literature. 
We show the token lengths (GPT-4 tokenizer) of questions and answers for each task in Figure~\ref{fig:length}.
SciAssess also includes five types of questions (\ie true/false questions, multiple-choice questions, table extraction, text extraction, and molecule generation) with four metrics (\ie accuracy, recall, F1-score, and molecule similarity).
For detailed descriptions and concrete examples, please refer to Appendix \ref{app:question-type}.
We also provide general prompt template and specific prompt for each task in Appendix \ref{app:prompt-template} and \ref{app:task-prompt}, respectively.

\subsubsection{Biology} 
Biological literature encompasses a wealth of specialized terminology and complex concepts, as well as a significant amount of non-textual information such as tables and figures.
Effectively extracting and integrating these elements presents a crucial challenge. 
Given that tasks in the biological domain typically require precise identification and understanding of intricate biological entities, processes, and relationships, we have selected a set of representative tasks, including the recognition of specialized terminology, the comprehension of chart information, and the extraction of entity relationships, to evaluate the performance in this field.

In this domain, following tasks are devised: MMLU-Pro-Biology, biology chart QA, chemical entities recognition, compound disease recognition, disease entities recognition, and gene disease function.
Detailed descriptions and prompts are provided in Appendix \ref{app:prompt-biology}.

\subsubsection{Chemistry}
The field of chemistry involves a vast array of complex molecular structures, chemical reactions, and properties, alongside a substantial amount of data presented in formulas, reaction equations, and diagrams. 
Effectively processing and interpreting these components is a significant challenge for language models. 
Tasks in the chemical domain demand precise understanding of molecular compositions, reaction mechanisms, and material properties. 
To evaluate the performance of LLMs in this domain, we have selected representative tasks such as the recognition of chemical compounds, the interpretation of reaction pathways, and the extraction of relationships between chemical entities.

We devise following tasks for organic materials: MMLU-Pro-Chemistry, electrolyte table QA, OLED property extraction, polymer chart QA, polymer composition extraction, polymer property extraction, solubility extraction, reactant QA, and reaction mechanism QA.
Detailed descriptions and prompt templates are provided in Appendix \ref{app:prompt-chemistry}.

\subsubsection{Materials}
Materials science encompasses a broad range of substances, including metals, ceramics, polymers, and composites, each with distinct properties and applications. 
These materials are widely used across industries such as aerospace, automotive, and construction. 
By fine-tuning their composition, structure, and processing techniques, materials can be engineered to meet specific performance requirements \cite{caron1983improvement}. 
Accurately extracting material compositions, structural characteristics, and process parameters from the literature is essential for advancing material design and optimization.

Specifically, following tasks are devised: material QA, Alloy Chart QA, composition extraction, temperature QA, sample differentiation, and treatment sequence.
Detailed descriptions and prompt templates are provided in Appendix \ref{app:prompt-material}.

\subsubsection{Medicine}
Medicine focuses on developing new therapeutics. 
Leveraging advanced intelligent tools, especially LLMs, can significantly enhance the efficiency and effectiveness of discovering and developing new drugs. 
To evaluate the capability of LLMs in this domain, it is imperative to develop specialized tasks that reflect the complexities and nuances of biomedical research.
By designing targeted tasks, we can better assess the ability of LLMs to navigate and interpret the wealth of information critical to the development of new therapeutics.

Specifically, we devise: MMLU-Pro-Health, affinity extraction, drug chart QA, tag to molecule, markush to molecule, molecule in document.
Detailed descriptions and prompts are provided in Appendix \ref{app:prompt-medicine}.

\subsection{Data Quality, Privacy, and Copyright}\label{sec:data_compliance}
To safeguard the quality and ethical standards, meticulous steps were undertaken in its preparation and validation:

\textbf{Distractor Construction:} 
Our data points are human-annotated, as well as the distractors. And how the distractors are determined depends on the specific task. 
For example, in value-type multiple-choice questions, the distractors are values near the ground truth. For extraction tasks, the distractors are other targets in the given context, except for the ground truth target.

\textbf{Expert Validation:} 
Each data point (as indicated by black tasks in Table \ref{tab:benchmark-stat}) is independently labeled by two annotators who are domain experts in the relevant fields.
If their labels agree, the label is accepted; if not, they engage in a discussion to determine the final label.
Their initial annotations have a Cohen's Kappa value~\cite{kappa} of 0.75, which indicates high reliability or agreement. 

\textbf{Screening and Anonymization:} 
Our annotators were instructed not to use any data samples containing sensitive information when building the benchmark.
For example, data samples including personal health information or specific drug details were carefully reviewed. 
If such sensitive information was identified, it was either anonymized by removing personal identifiers or replacing specific details with general terms, or the entire sample was excluded from the benchmark.

\textbf{Copyright Compliance:} 
Our benchmark includes two types of data: some are adopted from existing benchmarks, and others are constructed from scratch by our team. 
For the data adopted from existing benchmarks, we provide the corresponding sources. 
For the data we created, we have obtained the necessary copyrights for the files used. 
To ensure full compliance with copyright laws, our repository only provides the Digital Object Identifier (DOI) for papers or patent number, and does not distribute the actual documents.
Researchers need to download the necessary files independently. 
Detailed instructions is included in the codebase to guide researchers on how to place the downloaded documents into the designated folder.



    

%% file: chapters/3_experiment.tex
\section{Experiment}
\subsection{Experiment Setup} 
\begin{table*}[t]
\centering
\tiny
\setlength{\tabcolsep}{2pt}
\begin{tabular}{>{
\centering\arraybackslash}m{1.25cm} | >{
\centering\arraybackslash}p{3cm} | >{
\centering\arraybackslash}p{1.00cm} >{
\centering\arraybackslash}p{0.85cm} >{
\centering\arraybackslash}p{0.85cm} >{
\centering\arraybackslash}p{0.85cm} >{
\centering\arraybackslash}p{0.85cm} >{
\centering\arraybackslash}p{0.85cm} >{
\centering\arraybackslash}p{0.85cm} >{
\centering\arraybackslash}p{0.85cm} | >{
\centering\arraybackslash}p{0.85cm} >{
\centering\arraybackslash}p{0.85cm} >{
\centering\arraybackslash}p{0.85cm}
}
\toprule
Domain & Task & o1 & GPT-4o & GPT-4 & GPT-3.5 & Moonshot & Claude3 & Doubao & Gemini & Llama3.1 & Qwen2.5 & Mixtral \\
\noalign{\vskip -1pt}\midrule\noalign{\vskip -1pt}
\multirow{6}{1.2cm}{\centering Biology}
& MMLU-Pro-Biology* & \textcolor{orange}{\textbf{0.901}} & 0.874 & 0.845 & 0.650 & 0.755 & 0.781 & 0.770 & 0.842 & 0.815 & \textcolor{teal}{\textbf{0.840}} & 0.743 \\
& Biology Chart QA & \textcolor{orange}{\textbf{0.653}} & 0.558 & 0.442 & 0.312 & 0.518 & 0.402 & 0.523 & 0.548 & 0.477 & \textcolor{teal}{\textbf{0.487}} & 0.422 \\
& Chemical Entities Recognition* & \textcolor{orange}{\textbf{0.862}} & 0.795 & 0.817 & 0.649 & 0.821 & 0.815 & 0.749 & 0.745 & \textcolor{teal}{\textbf{0.836}} & 0.764 & 0.707 \\
& Compound Disease Recognition* & 0.745 & 0.733 & \textcolor{orange}{\textbf{0.753}} & 0.636 & 0.745 & 0.737 & 0.733 & 0.751 & \textcolor{teal}{\textbf{0.768}} & 0.712 & 0.757 \\
& Disease Entities Recognition* & \textcolor{orange}{\textbf{0.831}} & 0.763 & 0.670 & 0.688 & 0.654 & 0.684 & 0.742 & 0.767 & \textcolor{teal}{\textbf{0.793}} & \textcolor{teal}{\textbf{0.793}} & 0.737 \\
& Gene Disease Function* & \textcolor{orange}{\textbf{0.687}} & 0.410 & 0.587 & 0.391 & 0.538 & 0.506 & 0.539 & 0.558 & \textcolor{teal}{\textbf{0.474}} & 0.438 & 0.418 \\
\noalign{\vskip -1pt}\midrule\noalign{\vskip -1pt}
\multirow{9}{1.2cm}{\centering Chemistry} 
& MMLU-Pro-Chemistry* & \textcolor{orange}{\textbf{0.868}} & 0.745 & 0.621 & 0.303 & 0.428 & 0.496 & 0.446 & 0.683 & 0.676 & \textcolor{teal}{\textbf{0.723}} & 0.501 \\
& Electrolyte Table QA & \textcolor{orange}{\textbf{0.925}} & 0.855 & 0.810 & 0.305 & 0.765 & 0.435 & 0.745 & 0.765 & 0.755 & \textcolor{teal}{\textbf{0.785}} & 0.455 \\
& OLED Property Extraction & 0.394 & 0.438 & \textcolor{orange}{\textbf{0.455}} & 0.280 & 0.160 & 0.055 & 0.413 & 0.419 & \textcolor{teal}{\textbf{0.563}} & 0.499 & 0.355 \\
& Polymer Chart QA & \textcolor{orange}{\textbf{1.000}} & 0.867 & 0.733 & 0.667 & 0.800 & 0.467 & 0.400 & 0.733 & \textcolor{teal}{\textbf{0.800}} & \textcolor{teal}{\textbf{0.800}} & \textcolor{teal}{\textbf{0.800}} \\
& Polymer Composition QA & \textcolor{orange}{\textbf{0.986}} & 0.938 & 0.947 & 0.330 & 0.971 & 0.555 & 0.804 & 0.947 & 0.852 & \textcolor{teal}{\textbf{0.914}} & 0.493 \\
& Polymer Property Extraction & 0.606 & \textcolor{orange}{\textbf{0.759}} & 0.758 & 0.562 & 0.736 & 0.634 & 0.508 & 0.580 & 0.690 & \textcolor{teal}{\textbf{0.692}} & 0.573 \\
& Solubility Extraction & 0.427 & 0.444 & 0.431 & 0.408 & \textcolor{orange}{\textbf{0.445}} & 0.375 & 0.409 & 0.440 & \textcolor{teal}{\textbf{0.447}} & 0.437 & 0.314 \\
& Reactant QA & \textcolor{orange}{\textbf{0.559}} & 0.487 & 0.441 & 0.272 & 0.415 & 0.241 & 0.272 & 0.344 & \textcolor{teal}{\textbf{0.385}} & 0.379 & 0.231 \\
& Reaction Mechanism QA & \textcolor{orange}{\textbf{0.682}} & 0.591 & 0.273 & 0.227 & 0.500 & 0.136 & 0.409 & \textcolor{orange}{\textbf{0.682}} & 0.455 & \textcolor{teal}{\textbf{0.500}} & 0.273 \\
\noalign{\vskip -1pt}\midrule\noalign{\vskip -1pt}
\multirow{6}{1.2cm}{\centering Material} 
& Material QA & \textcolor{orange}{\textbf{0.821}} & 0.768 & 0.722 & 0.521 & 0.620 & 0.620 & 0.669 & 0.722 & \textcolor{teal}{\textbf{0.738}} & 0.719 & 0.631 \\
& Alloy Chart QA & 0.533 & 0.467 & 0.200 & 0.400 & 0.333 & 0.467 & 0.533 & \textcolor{orange}{\textbf{0.600}} & 0.467 & \textcolor{teal}{\textbf{0.533}} & 0.467 \\
& Composition Extraction & \textcolor{orange}{\textbf{0.488}} & 0.462 & 0.467 & 0.189 & 0.423 & 0.427 & 0.398 & 0.389 & \textcolor{teal}{\textbf{0.457}} & 0.430 & 0.177 \\
& Temperature QA & 0.836 & 0.807 & 0.734 & 0.295 & \textcolor{orange}{\textbf{0.845}} & 0.353 & 0.488 & 0.715 & \textcolor{teal}{\textbf{0.652}} & 0.647 & 0.382 \\
& Sample Differentiation & 0.392 & 0.624 & 0.595 & 0.329 & \textcolor{orange}{\textbf{0.688}} & 0.245 & 0.376 & 0.586 & \textcolor{teal}{\textbf{0.624}} & 0.578 & 0.426 \\
& Treatment Sequence & 0.624 & 0.594 & 0.678 & 0.485 & \textcolor{orange}{\textbf{0.683}} & 0.480 & 0.658 & 0.634 & 0.614 & 0.658 & \textcolor{teal}{\textbf{0.673}} \\
\noalign{\vskip -1pt}\midrule\noalign{\vskip -1pt}
\multirow{6}{1.2cm}{\centering Medicine} 
& MMLU-Pro-Health* & \textcolor{orange}{\textbf{0.784}} & 0.763 & 0.715 & 0.531 & 0.644 & 0.614 & 0.605 & 0.663 & \textcolor{teal}{\textbf{0.710}} & 0.685 & 0.603 \\
& Affinity Extraction & 0.068 & \textcolor{orange}{\textbf{0.101}} & 0.076 & 0.055 & 0.063 & 0.045 & 0.081 & 0.052 & 0.047 & \textcolor{teal}{\textbf{0.071}} & 0.049 \\
& Drug Chart QA & \textcolor{orange}{\textbf{0.600}} & 0.467 & 0.333 & 0.333 & 0.333 & 0.467 & 0.400 & 0.533 & \textcolor{teal}{\textbf{0.400}} & 0.333 & \textcolor{teal}{\textbf{0.400}} \\
& Tag2Mol & 0.127 & \textcolor{orange}{\textbf{0.229}} & 0.092 & 0.023 & 0.133 & 0.061 & 0.105 & 0.211 & \textcolor{teal}{\textbf{0.143}} & 0.136 & 0.021 \\
& Markush2Mol & 0.662 & 0.585 & \textcolor{orange}{\textbf{0.684}} & 0.523 & 0.391 & 0.503 & 0.565 & 0.683 & 0.425 & 0.443 & \textcolor{teal}{\textbf{0.576}} \\
& Mol In Document & \textcolor{orange}{\textbf{0.840}} & 0.600 & 0.620 & 0.440 & 0.480 & 0.640 & 0.560 & 0.580 & 0.520 & \textcolor{teal}{\textbf{0.600}} & 0.460 \\
\bottomrule
\end{tabular}
\caption{Performance comparison of LLMs across various scientific domains. 
\textcolor{orange}{\textbf{Orange}} and \textcolor{teal}{\textbf{green}} indicate the best in closed and open source LLMs, respectively.
Chain-of-thought prompt is implemented for each task and model, except for OpenAI-o1.
* indicates 3-shot.}
\label{tab:all_result}
\end{table*}
\textbf{Baseline LLMs.} 
To measure how leading LLMs perform on SciAssess, we benchmark extensively. 
For closed-source LLMs, we test OpenAI-o1, GPT-4o, GPT-4, GPT-3.5, Gemini-1.5-Pro, Claude 3 Opus, Moonshot-v1 and Doubao.
For open-source LLMs, we test Llama-3.1-70B, Mixtral-8x22B-Instruct-v0.1, and Qwen-2.5-72B. 
Briefs about all models are provided in Appendix \ref{app:baseline_model}.



\textbf{Experiment Workflow.} 
For closed-source models, we utilize the official API calls provided by the model developers, while for open-source models, we obtain these models from HuggingFace \cite{hf}, deploy them locally with vllm~\cite{vllm}, and then perform the tests. 
Tasks curated by us require real context from papers, thus the PDF content needs to be converted to text as inputs for LLMs. 
If the LLM includes a built-in PDF parsing interface (\eg Gemini and Moonshot), we simply use the interface; otherwise, we employ PyPDF2~\footnote{\url{https://pypdf2.readthedocs.io/en/3.x/}}, a widely-used open-source PDF parsing tool.
Our aim is to explore the ability boundary of LLMs, thus strategies that enhance LLMs' inference ability (\ie in-context learning~\cite{gpt3} and chain-of-thought~\cite{cot}) are adopted.
Specifically, due to the input length limitations of the LLMs, tasks requiring long context of a PDF document are executed in a zero-shot manner. 
Tasks that do not require such long context (\eg MMLU-Pro, entities recognition) are evaluated using 3-shot settings.
And chain-of-thought prompt is implemented in every task by prompting the model to think step-by-step before concluding.
The only exception is OpenAI-o1, whose official prompt guideline suggests users to ``avoid chain-of-thought prompts''.
We also provide complete performance evaluation results with chain-of-thought prompts in Appendix~\ref{app:performance_wo_cot}.

\subsection{Results and Analysis}
In this section, we analyze the performance of LLMs on SciAssess. 
The overall performance comparison, as summarized in Table \ref{tab:all_result}, reveals the distinct strengths and weaknesses of each model in science literature analysis. 

\subsubsection{Performances of Different Ability Levels}

\begin{table*}[t]
\centering
\tiny
\setlength{\tabcolsep}{0.75pt}
\begin{tabular}{>{
\centering\arraybackslash}m{2.5cm} | >{
\centering\arraybackslash}p{1.5cm}  >{
\centering\arraybackslash}p{1.5cm} | *{8}{>{
\centering\arraybackslash}p{0.85cm}} | *{3}{>{
\centering\arraybackslash}p{0.85cm}}
}
\toprule
Ability Level & Question Type & Metric & o1 & GPT-4o & GPT-4 & GPT-3.5 & Moonshot & Claude3 & Doubao & Gemini & Llama3.1 & Qwen2.5 & Mixtral \\
\noalign{\vskip -1pt}\midrule\noalign{\vskip -1pt}
 \multirow{2}{2.5cm}{\centering Memorization (L1)} & Multiple Choice & Accuracy & \textcolor{orange}{\textbf{0.843}} & 0.788 & 0.726 & 0.501 & 0.612 & 0.628 & 0.622 & 0.728 & 0.735 & \textcolor{teal}{\textbf{0.742}} & 0.619 \\ 
 \noalign{\vskip 2pt} \cline{2-14} \noalign{\vskip 2pt} 
 & \multicolumn{2}{c|}{Average Rank} & \textcolor{orange}{\textbf{1.000}} & 2.000 & 6.000 & 11.000 & 10.000 & 7.000 & 8.000 & 5.000 & 4.000 & \textcolor{teal}{\textbf{3.000}} & 9.000 \\ 
 \noalign{\vskip -1pt}\midrule\noalign{\vskip -1pt} 
 \multirow{5}{2.5cm}{\centering Comprehension (L2)} & Multiple Choice & Accuracy & \textcolor{orange}{\textbf{0.790}} & 0.708 & 0.600 & 0.377 & 0.652 & 0.449 & 0.556 & 0.692 & 0.629 & \textcolor{teal}{\textbf{0.643}} & 0.488 \\ 
 & Table Extraction & Recall & 0.397 & \textcolor{orange}{\textbf{0.441}} & 0.437 & 0.299 & 0.365 & 0.307 & 0.362 & 0.376 & \textcolor{teal}{\textbf{0.441}} & 0.426 & 0.294 \\ 
 & Text Extraction & F1-score & \textcolor{orange}{\textbf{0.781}} & 0.675 & 0.707 & 0.591 & 0.690 & 0.686 & 0.691 & 0.705 & \textcolor{teal}{\textbf{0.718}} & 0.677 & 0.655 \\ 
 & Mol. Generation & Mol. Similarity & 0.127 & \textcolor{orange}{\textbf{0.229}} & 0.092 & 0.023 & 0.133 & 0.061 & 0.105 & 0.211 & \textcolor{teal}{\textbf{0.143}} & 0.136 & 0.021 \\ 
 \noalign{\vskip 2pt} \cline{2-14} \noalign{\vskip 2pt} 
 & \multicolumn{2}{c|}{Average Rank} & \textcolor{orange}{\textbf{3.250}} & 3.375 & 5.250 & 10.500 & 5.500 & 8.750 & 7.000 & 3.750 & \textcolor{teal}{\textbf{3.125}} & 5.250 & 10.250 \\ 
 \noalign{\vskip -1pt}\midrule\noalign{\vskip -1pt} 
 \multirow{4}{2.5cm}{\centering Analysis \& Reasoning (L3)} & Multiple Choice & Accuracy & 0.544 & \textcolor{orange}{\textbf{0.567}} & 0.436 & 0.276 & 0.534 & 0.207 & 0.352 & 0.537 & \textcolor{teal}{\textbf{0.488}} & 0.486 & 0.310 \\ 
 & Mol. Generation & Mol. Similarity & 0.662 & 0.585 & \textcolor{orange}{\textbf{0.684}} & 0.523 & 0.391 & 0.503 & 0.565 & 0.683 & 0.425 & 0.443 & \textcolor{teal}{\textbf{0.576}} \\ 
 & True/False & Accuracy & \textcolor{orange}{\textbf{0.732}} & 0.597 & 0.649 & 0.462 & 0.582 & 0.560 & 0.609 & 0.607 & 0.567 & \textcolor{teal}{\textbf{0.629}} & 0.566 \\ 
 \noalign{\vskip 2pt} \cline{2-14} \noalign{\vskip 2pt} 
 & \multicolumn{2}{c|}{Average Rank} & \textcolor{orange}{\textbf{2.000}} & 3.667 & 3.333 & 9.333 & 7.333 & 9.667 & 6.000 & 3.333 & 7.667 & \textcolor{teal}{\textbf{6.000}} & 7.667 \\ 
\bottomrule
\end{tabular}
\caption{Performance on Memorization (L1), Comprehension (L2), and Analysis \& Reasoning (L3) tasks.}
\label{tab:result_abilities}
\end{table*}

\begin{figure*}[t]
    \centering
    \includegraphics[width=\textwidth]{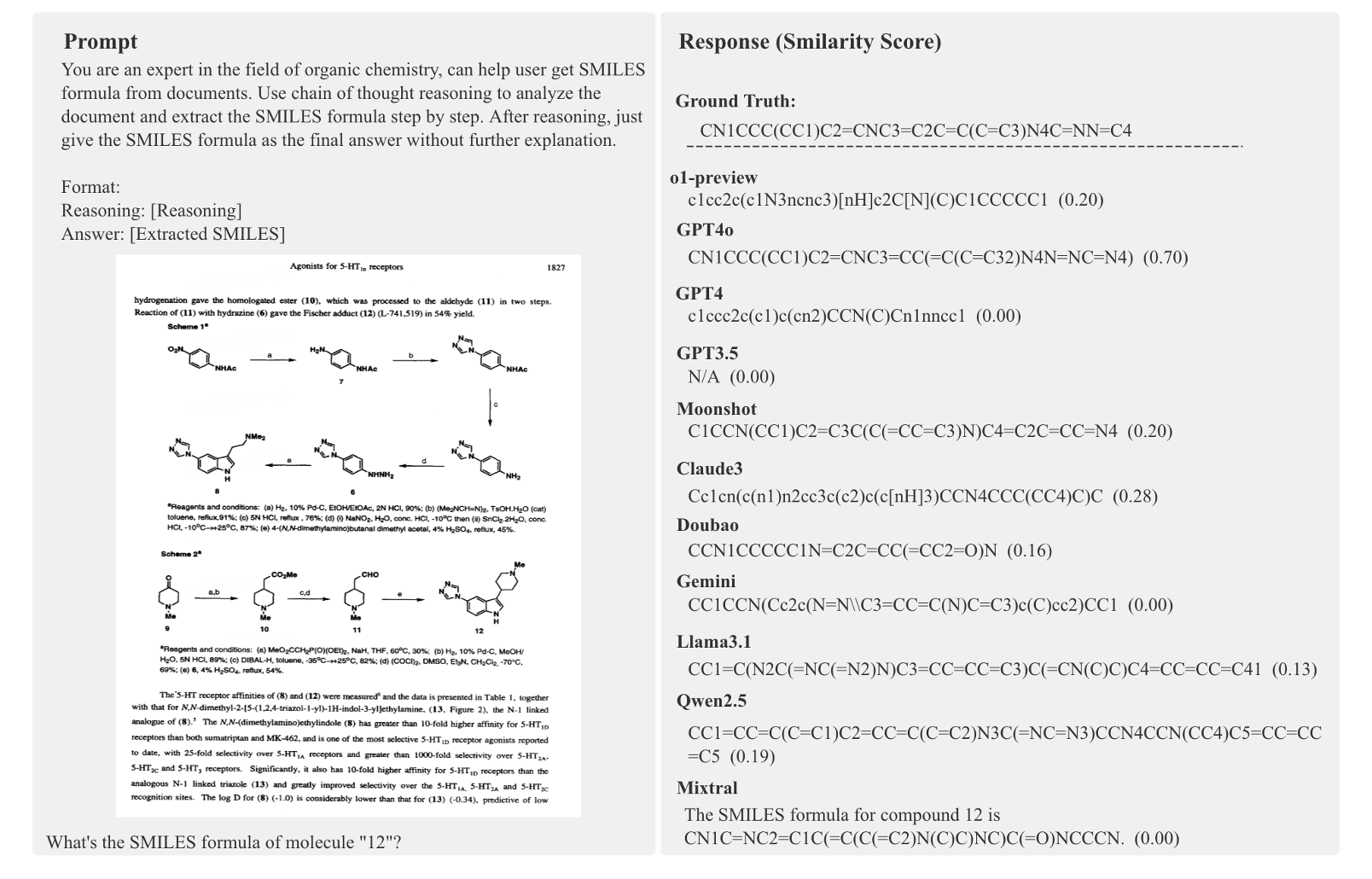}
    \vspace{-14pt}
    \caption{Example of Tag to Molecule task.}
    \label{fig:tag2mol}
\end{figure*}

Table \ref{tab:result_abilities} presents the performance of evaluated LLMs across three progressive ability level.
Tasks are further categorized according to their question types, with average results and rankings provided for each ability levels.
We observe the following:
(1) \textbf{Memorization (L1):} 
OpenAI-o1 and Qwen-2.5 demonstrates the highest average accuracy of 0.843 and 0.742, respectively, indicating consistently superior performance in memorization tasks. 
(2) \textbf{Comprehension (L2):} 
OpenAI-o1 excels in multiple-choice and text extraction comprehension with accuracy of 0.790 and 0.781, respectively, and maintains the top average rank of 3.25.
Notably, the only L2-level molecule generation task, \textit{Tag to Molecule}, reveals poor performance across all LLMs.
As illustrated in Figure \ref{fig:tag2mol}, current PDF parsing technologies, whether open-source like PyPDF or proprietary like Gemini or Moonshot, fail to effectively parse molecular structures in documents. 
Consequently, LLMs struggle with the \textit{Tag to Molecule} task. 
We propose that a critical advancement for future LLM-based literature understanding assistant is the integration of PDF parsing solutions capable of recognizing molecular structures.
(3) \textbf{Analysis \& Reasoning (L3):}
The average rank reveals OpenAI-o1, GPT-4, and Gemini are the top performers with ranks of 2.00, 3.33, and 3.33, respectively.

Overall, OpenAI-o1 consistently ranks high across all ability levels.
GPT-4o and Gemini also demonstrate strong overall performance, especially in memorization and reasoning.

Based on these observations, we suggest the following recommendations: (1) For tasks heavily reliant on memorization, OpenAI-o1 and Qwen2.5 are recommended due to their high accuracy and ranking;
(2) For comprehension tasks, particularly those involving complex data extraction and generation, OpenAI-o1 is an ideal choice.
(3) For analysis and reasoning tasks, OpenAI flagship models (\ie o1, GPT4, GPT-4o), and Gemini provide reliable performance and should be considered.

\subsubsection{Performance on Multimodal Contents}

\begin{table*}[t]
\centering
\tiny
\setlength{\tabcolsep}{2pt}
\begin{tabular}{>{
\centering\arraybackslash}m{1.25cm} | >{
\centering\arraybackslash}p{2cm} >{
\centering\arraybackslash}p{1.75cm} | *{8}{>{
\centering\arraybackslash}p{0.85cm}} | *{3}{>{
\centering\arraybackslash}p{0.85cm}}
}
\toprule
Modality & Question Type & Metric & o1 & GPT-4o & GPT-4 & GPT-3.5 & Moonshot & Claude3 & Doubao & Gemini & Llama3.1 & Qwen2.5 & Mixtral \\
\noalign{\vskip -1pt}\midrule\noalign{\vskip -1pt}
\multirow{3}{1.25cm}{\centering Text Only}
& Multiple Choice & Accuracy & \textcolor{orange}{\textbf{0.798}} & 0.788 & 0.740 & 0.423 & 0.707 & 0.523 & 0.594 & 0.737 & 0.724 & \textcolor{teal}{\textbf{0.729}} & 0.540 \\ 
& Text Extraction & F1-score & \textcolor{orange}{\textbf{0.781}} & 0.675 & 0.707 & 0.591 & 0.690 & 0.686 & 0.691 & 0.705 & \textcolor{teal}{\textbf{0.718}} & 0.677 & 0.655 \\ 
& True/False & Accuracy & 0.624 & 0.594 & 0.678 & 0.485 & \textcolor{orange}{\textbf{0.683}} & 0.480 & 0.658 & 0.634 & 0.614 & 0.658 & \textcolor{teal}{\textbf{0.673}} \\ 
\noalign{\vskip 1.5pt} \cline{2-14} \noalign{\vskip 1.5pt} 
& \multicolumn{2}{c|}{Average Rank} & 3.000 & 6.667 & \textcolor{orange}{\textbf{2.667}} & 10.667 & 4.667 & 9.333 & 5.833 & 4.667 & \textcolor{teal}{\textbf{5.333}} & 5.833 & 7.333 \\ 
\noalign{\vskip -1pt}\midrule\noalign{\vskip -1pt} 
\multirow{1}{1.25cm}{\centering Chart}
& Multiple Choice & Accuracy & \textcolor{orange}{\textbf{0.696}} & 0.590 & 0.427 & 0.428 & 0.496 & 0.451 & 0.464 & 0.604 & 0.536 & \textcolor{teal}{\textbf{0.538}} & 0.522 \\ 
\noalign{\vskip 1.5pt} \cline{2-14} \noalign{\vskip 1.5pt} 
& \multicolumn{2}{c|}{Average Rank} & \textcolor{orange}{\textbf{1.000}} & 3.000 & 11.000 & 10.000 & 7.000 & 9.000 & 8.000 & 2.000 & 5.000 & \textcolor{teal}{\textbf{4.000}} & 6.000 \\ 
\noalign{\vskip -1pt}\midrule\noalign{\vskip -1pt} 
\multirow{1}{1.25cm}{\centering Reaction}
& Multiple Choice & Accuracy & \textcolor{orange}{\textbf{0.620}} & 0.539 & 0.357 & 0.250 & 0.458 & 0.188 & 0.340 & 0.513 & 0.420 & \textcolor{teal}{\textbf{0.440}} & 0.252 \\ 
\noalign{\vskip 1.5pt} \cline{2-14} \noalign{\vskip 1.5pt} 
& \multicolumn{2}{c|}{Average Rank} & \textcolor{orange}{\textbf{1.000}} & 2.000 & 7.000 & 10.000 & 4.000 & 11.000 & 8.000 & 3.000 & 6.000 & \textcolor{teal}{\textbf{5.000}} & 9.000 \\ 
\noalign{\vskip -1pt}\midrule\noalign{\vskip -1pt} 
\multirow{3}{1.25cm}{\centering Mol.}
& Table Extraction & Recall & 0.231 & \textcolor{orange}{\textbf{0.270}} & 0.266 & 0.168 & 0.112 & 0.050 & 0.247 & 0.236 & \textcolor{teal}{\textbf{0.305}} & 0.285 & 0.202 \\ 
& Mol. Generation & Mol. Similarity & 0.394 & 0.407 & 0.388 & 0.273 & 0.262 & 0.282 & 0.335 & \textcolor{orange}{\textbf{0.447}} & 0.284 & 0.290 & \textcolor{teal}{\textbf{0.298}} \\ 
& True/False & Accuracy & \textcolor{orange}{\textbf{0.840}} & 0.600 & 0.620 & 0.440 & 0.480 & 0.640 & 0.560 & 0.580 & 0.520 & \textcolor{teal}{\textbf{0.600}} & 0.460 \\ 
\noalign{\vskip 1.5pt} \cline{2-14} \noalign{\vskip 1.5pt} 
& \multicolumn{2}{c|}{Average Rank} & 3.667 & \textcolor{orange}{\textbf{3.167}} & 3.667 & 10.000 & 10.000 & 7.333 & 5.667 & 4.333 & 5.667 & \textcolor{teal}{\textbf{4.500}} & 8.000 \\ 
\noalign{\vskip -1pt}\midrule\noalign{\vskip -1pt} 
\multirow{2}{1.25cm}{\centering Table}
& Multiple Choice & Accuracy & \textcolor{orange}{\textbf{0.925}} & 0.855 & 0.810 & 0.305 & 0.765 & 0.435 & 0.745 & 0.765 & 0.755 & \textcolor{teal}{\textbf{0.785}} & 0.455 \\ 
& Table Extraction & Recall & 0.397 & \textcolor{orange}{\textbf{0.441}} & 0.437 & 0.299 & 0.365 & 0.307 & 0.362 & 0.376 & \textcolor{teal}{\textbf{0.441}} & 0.426 & 0.294 \\ 
\noalign{\vskip 1.5pt} \cline{2-14} \noalign{\vskip 1.5pt} 
& \multicolumn{2}{c|}{Average Rank} & 3.000 & \textcolor{orange}{\textbf{1.750}} & 3.000 & 10.500 & 6.250 & 9.500 & 8.000 & 5.750 & 4.250 & \textcolor{teal}{\textbf{4.000}} & 10.000 \\ 
\bottomrule
\end{tabular}
\caption{Performance on multimodal contents.}
\label{tab:result_multimodal}
\end{table*}

Table \ref{tab:result_multimodal} summarizes the performance of LLMs on multimodal content tasks. 
For each modality, performances are averaged over different question types.
We observe the following:
(1) \textbf{Text-only tasks:} GPT-4 achieves the highest average rank (2.00).
(2) \textbf{Chart tasks:} OpenAI-o1 exhibit the highest accuracy (0.696).
(3) \textbf{Chemical reaction tasks:} OpenAI-o1 stands out with high accuracy in multiple-choice questions (0.62).
(3) \textbf{Molecule tasks:} GPT-4o excels with average ranks of 3.17, particularly in table extraction task.
(5) \textbf{Table tasks:} GPT-4o lead with the highest table extraction recall (0.44).

Overall, OpenAI models consistently rank as top performers across most modalities. 
Gemini also demonstrate strong performance, especially in molecule generation tasks. 

Based on these observations, we suggest the following recommendations: 
(1) For text-only tasks, OpenAI-o1 and GPT-4o are highly recommended due to their superior accuracy and ranking.
(2) For chart and chemical reaction tasks, OpenAI-o1 excels, making it suitable for such specialized applications.
(3) For molecule structure and tabular tasks, GPT-4o is the preferred model, given it remarkable performance.

%% file: chapters/4_related_work.tex
\section{Related Work}
\label{sec:related_work}
\textbf{General benchmarks for LLMs.}
LLMs are evaluated across a variety of benchmarks to comprehensively assess their capabilities. Some benchmarks, such as MMLU \cite{mmlu}, MMLU-pro \cite{mmlu-pro}, CMMLU \cite{cmmlu}, and Xiezhi \cite{xiezhi}, are instrumental in evaluating models' world knowledge across diverse domains. 
For reasoning capabilities, benchmarks like GSM8k \cite{gsm8k} and BBH \cite{bbh} provide rigorous assessments of models' problem-solving and logical reasoning skills. 
In the realm of programming, benchmarks such as HumanEval \cite{codex-eval} and MBPP \cite{mbpp} serve as popular testbeds for evaluating models' coding proficiency.  
Additionally, TruthfulQA \cite{truthful-qa} and HaluEval \cite{halu-eval} are pivotal in assessing the veracity of models' outputs, ensuring their alignment with factual information.

Although some general benchmarks include a subset of science subjects, they mostly focus on Memorization (L1) and often overlook higher-level abilities such as Comprehension (L2) and Analysis \& Reasoning (L3). 
Furthermore, these benchmarks lack context-involved tasks, for example, understanding and reasoning over a scientific paper.

\textbf{Scientific literature benchmarks.}
Prior works have made significant strides in developing LLM benchmarks to assess the understanding of scientific literature. 
In the biomedical domain, notable efforts include BLUE \cite{blue}, which provides a set of tasks for evaluating models on various aspects of biomedical text-mining. 
Building on this, BLURB \cite{blurb} offers an extensive collection of datasets to further refine model performance in this specialized field. 
More recently, InBoXBART \cite{In-BoXBART} has been introduced, focusing on integrating information across multiple biomedical documents. 
SciRIFF~\cite{sciriff} is designed to extract and synthesize information from research literature across various scientific disciplines.

Compared with existing scientific literature benchmarks, SciAssess focuses more on tasks for interpreting multi-modal content (\eg molecular structures and tables), which are common in scientific literature.
Moreover, it features a real-world application scenarios that LLMs digest parsed PDF contents with parsing errors.

%% file: chapters/5_conclusion.tex
\section{Conclusion and Future Work}
SciAssess rigorously assesses the capabilities of LLMs for scientific literature analysis. 
It focuses four specialized areas: biology, chemistry, material, and medicine. 
The benchmark focuses on assessing LLMs' core competencies in Memorization (L1), Comprehension (L2), and Analysis \& Reasoning (L3) within the context of scientific literature analysis.
Through detailed evaluations of 11 LLMs, SciAssess highlights their strengths and identifies areas needing improvement across various ability levels, content modalities, and contextual scenarios.
Additionally, we emphasize the urgent need for PDF parsing algorithms tailored to handle content of various modalities, such as molecular structures and chemical reactions.
We hope that SciAssess supports the ongoing development of LLMs in scientific literature analysis. 
Looking ahead, we plan to broaden the range of scientific domains included in SciAssess and incorporate more vertical domains.
These enhancements aim to improve the benchmark's utility and efficacy, providing clearer guidance and fostering the advancement of LLMs in scientific literature analysis.

\section*{Limitation}

While SciAssess provides a comprehensive and valuable benchmarking suite across four primary domains -- biology, chemistry, material, and medicine -- there are several limitations to consider. 
Firstly, the scope of SciAssess is currently constrained to these four domains, with potential future extensions to other vertical domains such as physics and engineering.

Secondly, the creation and curation of high-quality, domain-specific training data are essential for the effective evaluation and improvement of LLMs. 
However, due to the high cost associated with manual labeling, 
SciAssess does not provide additional training data for these tasks. 
This absence of supplementary data can limit the ability of researchers to fine-tune and enhance LLMs specifically for the tasks included in SciAssess. 
Consequently, the benchmark results might reflect the inherent capabilities of the models rather than their optimized performance for each specific domain.

Lastly, while SciAssess aims to provide a rigorous evaluation framework, the complexity and diversity of scientific domains present challenges in ensuring comprehensive coverage and fairness. 
Some tasks may inherently favor certain types of models or architectures, leading to potential biases in performance evaluation.

\section*{Broader Impact}
Our work on benchmarking scientific literature analysis aligns with the scope of existing LLM benchmarks such as MMLU-pro. 
This paper represents progress in calibrating LLMs for specific domains, thereby amplifying the impacts that LLM benchmarks have had (and will continue to have) on the broader world. 
Additionally, we have not identified any ethical concerns or potential risks associated with this work.

%% file: chapters/6_appendix.tex
\newpage
\appendix



\section{Question Type}
\label{app:question-type}

\begin{figure*}[t]
  \centering
  \includegraphics[width=\textwidth]{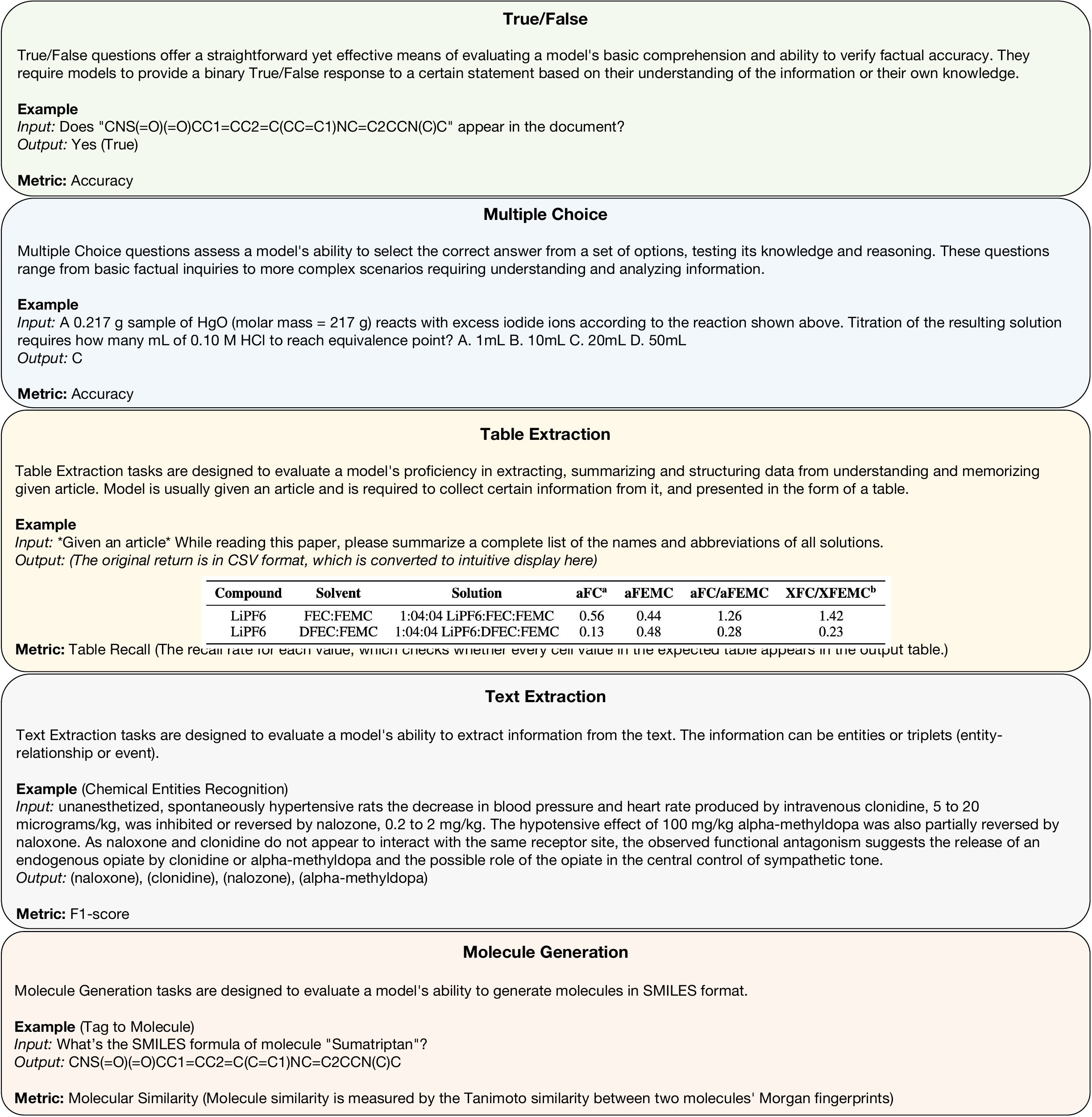}
  \vspace{-12pt}
  \caption{Question types.}
  \vspace{-12pt}
  \label{fig:question-type}
\end{figure*}

Five types of questions, as illustrated in Figure \ref{fig:question-type} are devised to evaluate the models. 
Each question type is accompanied by a detailed description and representative examples, along with the corresponding metrics used for assessment. 
For convenience, the input in each example is simplified, and its instruction is omitted.

\section{General Prompt Template}
\label{app:prompt-template}

We design following general prompt template for scientific literature analysis. 
It consists of: a system message defining the role of the assistant, the task description, some optional few-shot examples, and a user prompt of the question.

\begin{prompt}{Prompt Template}
\textbf{\small Role setting and task description:}

You are a highly intelligent assistant who answers the following multiple choice question correctly.
\\
\\
\textbf{\small Few-shot examples:}

Question: <question 1> \\
Answer: <answer 1> \\
... \\
Question: <question n> \\
Answer: <answer n>
\\
\\
\textbf{\small Question:}

Predict the number of lines in the EPR spectrum of a solution of 13C-labelled methyl radical (13CH3•), assuming the lines do not overlap.

a) 4

b) 3

c) 6

d) 24
\end{prompt}

\section{Task Prompt}
\label{app:task-prompt}

In this section, we detail the prompt templates for all tasks in SciAssess benchmark. We will introduce these templates in the following order: Biology (Section \ref{app:prompt-biology}), Chemistry (Section \ref{app:prompt-chemistry}), Material (Section \ref{app:prompt-material}) and Medicine (Section \ref{app:prompt-medicine}).

\subsection{Biology}
\label{app:prompt-biology}

\subsubsection{MMLU-Pro-Biology}
\label{app:prompt-mmlu-pro-biology}

\begin{prompt}{Prompt}
\textbf{\small System Message:}

You are a highly intelligent assistant who answers the following multiple choice question correctly.
Use chain of thought reasoning and provide reasoning before selecting the correct answer (e.g., a)xxx, or b)xxx).

Format:

Reasoning: [Reasoning]

Answer: [Answer]
\\
\\
\textbf{\small User Message:}

Which of the following would most likely provide examples of mitotic cell divisions?\\
a) cross section of muscle tissue\\
b) longitudinal section of a shoot tip\\
c) longitudinal section of a leaf vein\\
d) cross section of a fruit\\
e) cross section of a leaf\\
f) longitudinal section of a petal\\
g) longitudinal section of a seed\\
h) cross section of an anther (site of pollen production in a flower)
\\
\\
\textbf{\small Expected Answer:}

b) longitudinal section of a shoot tip
\end{prompt}

\subsubsection{Biology Chart QA}
\label{app:prompt-biology-chart-qa}
The analysis and understanding of biological properties, compositions, and processing techniques are critical for the discovery and development in life sciences. 
Often, this information is presented in charts, making it essential to extract and integrate such information with textual data. 
To assess the retrieval capabilities of models in the context of biological chart information, we have designed multiple-choice questions.

\begin{prompt}{Prompt}
\textbf{\small System Message:}

You are an expert in the field of Biomedical. You are a highly intelligent biology scientist who answers the following multiple choice question correctly.
Use chain of thought reasoning and provide reasoning before selecting the correct answer (e.g., a)xxx, or b)xxx).

Format:

Reasoning: [Reasoning]

Answer: [Answer]
\\
\\
\textbf{\small User Message:}

In Figure 3, which has a higher accurate score, with the graph encoder or without? 

a) with graph encoder\\
b) w/o graph encoder
\\
\\
\textbf{\small Expected Answer:}

a) with graph encoder
\end{prompt}

\subsubsection{Chemical Entities Recognition}
\label{app:prompt-chemical-entities-recognition}

This task involves recognizing chemical entity names using data from B5CDR \cite{CDR2016} and additional expert-annotated data. 
It evaluates the performance of LLMs in identifying complex drug names. The prompt template is as follows.

\begin{prompt}{Prompt}
\textbf{\small System Message:}

You are an expert in the field of Biomedical. I'll give you the abstract of literature. Please use chain of thought reasoning to identify all the compound entities in the abstract. First, analyze the abstract step by step, explaining your reasoning for identifying each compound entity. Then, provide a final list of the compound entities you recognized in the format: (compound 1), (compound 2), (compound 3).

Format:

Reasoning: [Reasoning]

Answer: [List of identified compounds]
\\
\\
\textbf{\small User Message:}

In unanesthetized, spontaneously hypertensive rats the decrease in blood pressure and heart rate produced by intravenous clonidine, 5 to 20 micrograms/kg, was inhibited or reversed by nalozone, 0.2 to 2 mg/kg. The hypotensive effect of 100 mg/kg alpha-methyldopa was also partially reversed by naloxone. Naloxone alone did not affect either blood pressure or heart rate. In brain membranes from spontaneously hypertensive rats clonidine, 10(-8) to 10(-5) M, did not influence stereoselective binding of [3H]-naloxone (8 nM), and naloxone, 10(-8) to 10(-4) M, did not influence clonidine-suppressible binding of [3H]-dihydroergocryptine (1 nM). These findings indicate that in spontaneously hypertensive rats the effects of central alpha-adrenoceptor stimulation involve activation of opiate receptors. <rest of the input>.
\\
\\
\textbf{\small Expected Answer:}

(naloxone), (clonidine), (nalozone), (alpha-methyldopa)
\end{prompt}

\subsubsection{Compound Disease Recognition}
\label{app:prompt-compound-disease-recognition}

Proposed in B5CDR \cite{CDR2016}, this task evaluates the capability of LLMs to identify and understand associations between compounds and diseases. Examples of process text:
\begin{prompt}{Example Paragraph}
Twenty children with acute lymphoblastic leukemia who developed meningeal disease were treated with a high-dose intravenous methotrexate regimen that was designed to achieve and maintain CSF methotrexate concentrations of 10(-5) mol/L without the need for concomitant intrathecal dosing. The methotrexate was administered as a loading dose of 6,000 mg/m2 for a period of one hour followed by an infusion of 1,200 mg/m2/h for 23 hours. Leucovorin rescue was initiated 12 hours after the end of the infusion with a loading dose of 200 mg/m2 followed by 12 mg/m2 every three hours for six doses and then every six hours until the plasma methotrexate level decreased to less than 1 X 10(-7) mol/L. The mean steady-state plasma and CSF methotrexate concentrations achieved were 1.1 X 10(-3) mol/L and 3.6 X 10(-5) mol/L, respectively. <rest of the paragraph>.
\end{prompt}

We then prompt the model with the following:

\begin{prompt}{Prompt}
\textbf{\small System Message:}

You are a biologist AI. I'll give you the abstract of literature. Please use chain of thought reasoning to identify all the (compound, disease) relations in the abstract. First, analyze the abstract step by step, explaining your reasoning for identifying each relation. Then, provide a final list of the relations in the format: '(compound 1, disease 1),(compound 2, disease 2),(compound 3, disease 3),....', without adding any additional comments or explanations.

Format:

Reasoning: [Reasoning]

Answer: [List of recognized relations]
\\
\\
\textbf{\small User Message:}

[processed text]
\\
\\
\textbf{\small Expected Answer:}

(methotrexate, transient hemiparesis), (methotrexate, neutropenia), (methotrexate, seizures), (methotrexate, mucositis)

\end{prompt}

\subsubsection{Disease Entities Recognition}
\label{app:prompt-disease-entities-recognition}

Similarly, this task involves recognizing disease entity names using data from \cite{CDR2016} and additional expert-annotated data, evaluating the performance of LLMs in identifying specialized disease names:

\begin{prompt}{Prompt}
\textbf{\small System Message:}

You are an expert in the field of Biomedical. You are a biologist AI. I'll give you the abstract of literature. Please use chain of thought reasoning to identify all the disease entities in the abstract. First, analyze the abstract step by step, explaining your reasoning for identifying each disease entity. Then, provide a final list of the disease entities you recognized in the format: (disease 1), (disease 2), (disease 3).

Format:

Reasoning: [Reasoning]

Answer: [List of recognized diseases]
\\
\\
\textbf{\small User Message:}

In unanesthetized, spontaneously hypertensive rats the decrease in blood pressure and heart rate produced by intravenous clonidine, 5 to 20 micrograms/kg, was inhibited or reversed by nalozone, 0.2 to 2 mg/kg. The hypotensive effect of 100 mg/kg alpha-methyldopa was also partially reversed by naloxone. Naloxone alone did not affect either blood pressure or heart rate. In brain membranes from spontaneously hypertensive rats clonidine, 10(-8) to 10(-5) M, did not influence stereoselective binding of [3H]-naloxone (8 nM), and naloxone, 10(-8) to 10(-4) M, did not influence clonidine-suppressible binding of [3H]-dihydroergocryptine (1 nM). These findings indicate that in spontaneously hypertensive rats the effects of central alpha-adrenoceptor stimulation involve activation of opiate receptors. <rest of the input>.
\\
\\
\textbf{\small Expected Answer:}

(hypertensive), (hypotensive)
\end{prompt}

\subsubsection{Gene Disease Function}
\label{app:prompt-gene-disease-function}
The Gene Disease Text Mining task focuses on "Gene-Disease" association semantics text mining. 
It evaluates the ability of models to extract and understand relationships between genes and diseases from scientific literature, with a focus on identifying gene and disease entities \cite{GDAS2022}. Examples of process text:

\begin{prompt}{Example Paragraph}
A novel frameshift mutation (+G) at codons 15/16 in a beta0 thalassaemia gene results in a significant reduction of beta globin mRNA values.

AIMS: To identify a novel beta globin gene mutation found in a Chinese family, and also to assess its functional consequences.

METHODS: Haematological analysis was performed on all family members. The 23 common mutations of beta thalassaemia found in Chinese populations were detected by means of a reverse dot blot method. Direct DNA sequencing of polymerase chain reaction (PCR) amplified complete beta globin gene was carried out to identify the novel mutation. A real time, one step reverse transcription PCR assay was used to measure beta globin mRNA in the reticulocytes of heterozygous patients.

RESULTS: A novel frameshift mutation-an insertion of G between codons 15 and 16 in a homonucleotide run of four guanines-was determined, which generates a new premature chain terminator at the 22nd codon. Relative quantitative analysis of the beta globin mRNA in heterozygous subjects demonstrated a 39.83\% reduction compared normal controls.

CONCLUSIONS: The significantly lower amounts of beta globin mRNA found in mutation carriers is probably caused by the rapid nonsense mediated degradation of the mutant mRNA. These data, combined with haematological analysis, suggest that this novel mutation of CDs 15/16 (+G) results in a beta(0) thalassaemia phenotype.
\end{prompt}

For extracting triplets (entities, semantic roles, entities), we prompt the model with:
\begin{prompt}{Prompt}
\textbf{\small System Message:}

You are an expert in the field of Biomedical. In this semantic role regconition task, you need to follow 3 steps, and finally just return me triples that needed. 

First, you need to identify the entities in the text. Entities can be classified into 2 categories--molecular, and trigger word. 'Molecular' includes disease, gene, protein, and enzyme. 'Trigger word' includes:

1)Variation(Var), which means DNA, RNA, and mutations in proteins and changes in molecular structure, e.g. 'mutations on the Arg248 and Arg282', 'mutant R282W', 'missense mutations'; 

2)Molecular Physiological Activity (MPA), including molecular activity, gene expression and molecular physiological activity, e.g. 'phosphorylation', 'transcription', 'histone methylation', 'bioactivation of cyclophosphamide'; 

3)Interaction, molecule-to-molecule or molecule-to-cell connections, e.g. 'bind', 'interaction'; 

4)Pathway, e.g. 'Bmp pathway','PI3K pathway'; 

5)Cell Physiological Activity (CPA), Activities at or above the cellular level, including cellular reactivity and cell or organ development and growth, e.g. 'T helper cell responses', 'renal development'; 

6)Regulation (Reg), a neutral cue word or phrase meaning no loss or gain, e.g. 'resolved in', 'regulated'; 

7)Positive Regulation (PosReg), a cue word or phrase that indicates the acquisition of a function, e.g. 'facilitates', 'enhanced', 'increased'; 

8)Negative Regulation (NegReg), a clue word or phrase that indicates a loss of function, e.g. 'suppressed', 'decreased', 'inhibited'. 

Second, you need to identify the semantic role labeling objects, including 'ThemeOf'(from the main thing entity to the current entity) and 'CauseOf'(From the current entity to the Cause entity). 

Third, please give me tripples that contain entities and semantic role labeling objects(ThemeOf or Causeof).

Use chain of thought reasoning to explain your process of identifying the entities and relations, and then provide the final triples in the format: (\ldots), (\ldots)

Format:

Reasoning: [Reasoning]

Answer: [List of recognized triples]
\\
\\
\textbf{\small User Message:}

[processed text]
\\
\\
\textbf{\small Expected Answer:}

(frameshift, CauseOf, reduction), (caused by, CauseOf, lower), (mutation, CauseOf, results in), (beta(0) thalassaemia, ThemeOf, results in), (beta globin mRNA, ThemeOf, reduction), (beta0 thalassaemia gene, ThemeOf, frameshift), (insertion, CauseOf, generates), (premature chain terminator, ThemeOf, generates), (amounts of beta globin mRNA, ThemeOf, lower), (mutation, CauseOf, caused by), (degradation, ThemeOf, caused by).
\end{prompt}

\subsection{Chemistry}
\label{app:prompt-chemistry}

\subsubsection{MMLU-Pro-Chemistry}
\label{app:prompt-mmlu-pro-chemistry}

The example of MMLU-Pro-Chemistry is similar to the MMLU-Pro-Biology task in Appendix \ref{app:prompt-mmlu-pro-biology}.








\subsubsection{Electrolyte Table QA}
\label{app:prompt-electrolyte-table-qa}

The composition and properties of organic electrolytes are crucial for battery performance, stability, and safety. To evaluate the model's retrieval capabilities regarding electrolyte information, we posed multiple-choice questions about the components of solution systems and the dissolution reactions, focusing on their physical and chemical properties as presented in the tables within the articles. We prompt the model with the following:
\begin{prompt}{Prompt}
\textbf{\small System Message:}

You are an expert in the electrolytes field. Please answer the following multiple choice question correctly.
Use chain of thought reasoning and provide reasoning before selecting the correct answer (e.g., a)xxx, or b)xxx).

Format:

Reasoning: [Reasoning]

Answer: [Answer]
\\
\\
\textbf{\small User Message:}

In the upper paper, what are the minimum and maximum intramolecular distances (nm) of dimethyl carbonate?

a) 0.41/0.87

b) 0.49/0.67

c) 0.25/0.25

d) 0.25/0.38
\\
\\
\textbf{\small Expected Answer:}

a) 0.41/0.87
\end{prompt}

\subsubsection{OLED Property Extraction}
\label{app:prompt-oled-property-extraction}

This task evaluates the LLM's ability to extract information about OLED molecules and their optical properties. 
It tests several key capabilities, including their understanding of complex and domain-specific language and their ability to interpret and extract data from tables. An example output is shown in Table \ref{tab:oled}.
\begin{table*}[t]
\tiny
\centering
\caption{OLED Property example.}
\begin{tabular}{ccccccc}
\toprule
Host & Dopant&Td [°C] / Tg [°C] / ET [eV] & Von [V] & max EQE [\%] /  CE [cd A-1] / PE [lm W-1] & EQE [\%] / CE [cd A-1] / PE [lm W-1]& CIE [x, y] \\
\midrule
PPO1 & FCNIr & – / 74 / 3.02 & – & 17.1 / 20.5 / 14.3 & – / – / – & (0.14, 0.16) \\
PPO2 & FCNIr & – /123 / 3.02 & – & 18.4 / 21.1 / 16.6 & – / – / – & (0.14, 0.15)  \\
mCPPO1 & FCNIrpic & – /– / 3.00 & – & 25.1 / – / 29.8 & 23.1 / 28.9 / 15.1 & (0.14, 0.18) \\
CDPO & 5CzCN & 455 / 89 / 2.84 & 4.9 & 13.2 / 31.6 / 18.1 & – / – / – & (0.20, 0.38) \\
\bottomrule
\end{tabular}
\label{tab:oled}
\end{table*}
We prompt the model with the following:

\begin{prompt}{Prompt}

\textbf{\small System Message:}

You are an expert in the field of organic photovoltics. Please give a complete list of Host, Host's SMILES structure (if exists), Dopant, Assistant Dopant (if exists), Td/Tg/ET, Von,max EQE/CE/PE,EQE/CE/PE, and CIE [x, y]

* Output in csv format with columns of those attributes, do not write units only the value like "10.5".

* Quote the column name or Host's Name or Dopant's Name if it contains space or special characters like ",".

* If there are multiple tables, concat them. Don't give me reference or using "...", give me complete table!

* Should return all columns mentioned, if empty just return `NaN`. "Host" and "Dopant" should not be empty.

* "Host" and "Dopant" should be short name of the organic molecule.

* Should find more information from the whole content, including tables, text.

for example, you should return:

```csv\\
Host,SMILES,Dopant,Td /Tg /ET, \\
Von,max EQE/CE/PE,EQE/CE/PE,"CIE"\\
PPO1,O=P(c1ccccc1)(c1ccccc1)c1ccccc1,FCNIr,–/74/3.02,–,\\
17.1/20.5/14.3,–/–/–,"(0.14, 0.16)"\\
PPO2,O=P(c1ccccc1)(c1ccccc1)c1ccccc1,FCNIr,–/123/3.02,–,\\
18.4/21.1/16.6,–/–/–,"(0.14, 0.15)"\\
```

Please use a step-by-step approach to analyze the content and ensure that all relevant information is accurately extracted. Only provide reasoning for how you identified each attribute and output the final csv format.

Format:

Reasoning: [Reasoning]

Answer: [Extracted csv]
\\
\\
\textbf{\small User Message:}

[ docoment.pdf ]

\end{prompt}

\subsubsection{Polymer Chart QA}
\label{app:prompt-polymer-chart-qa}

The processing steps and properties of polymer materials are often represented through charts. 
Extracting information from these charts and integrating it with textual data is crucial. 
To further assess the retrieval capabilities of models concerning polymer chart information, we designed multiple-choice questions involving polymer composition, processing techniques, and properties.

The example of Polymer Chart QA can be found in a similar format to the Biology Chart QA task in Appendix \ref{app:prompt-biology-chart-qa}.

\subsubsection{Polymer Composition QA}
\label{app:prompt-polymer-composition-qa}

This task involves extracting the blend ratio of donor to acceptor in the most efficient solar cell from the text of scientific literature. 

We prompt the model with the following:
\begin{prompt}{Prompt}
\textbf{\small System Message:}

You are an expert in the field of polymer solar cells researcher who answers the following multiple choice question correctly.
Use chain of thought reasoning and provide reasoning before selecting the correct answer (e.g., a)xxx, or b)xxx).

Format:

Reasoning: [Reasoning]

Answer: [Answer]
\\
\\
\textbf{\small User Message:}

In this paper, What is the blend ratio of donor to acceptor in the most efficient solar cell?  

a) 1:4 

b) 20:8 

c) 30:50 

d) 2:4
\\
\\
\textbf{\small Expected Answer:}

a) 1:4
\end{prompt}

\subsubsection{Polymer Property Extraction}
\label{app:prompt-polymer-property-extraction}

This task focuses on extracting vital values such as power conversion efficiency (PCE) and open-circuit voltage ($V_{\mathrm{OC}}$) from tables within the literature. 

We prompt the model with the following:
\begin{prompt}{Prompt}

\textbf{\small System Message:}

You are an expert in the field of polymer solar cells researcher.

Please give a complete list of Nickname, PCE\_max, PCE\_ave, Voc , Jsc, FF;
* Output in csv format with columns of those attribution, do not write units only the value like "10.5".

* If there are multiple tables, concat them. Don't give me reference or using "...", give me complete table!

* Should return all columns mentioned, if empty just return `NaN`. Nickname should not be empty.

* Nickname should be short name of polymers, for example: `PCBM:PffBT4T-2OD:PC61PM` should return `PffBT4T-2OD`.

* Only return acceptor `PC71BM` related records.

* If with different experiment settings for the same nickname, only return the record with `highest PCE` !

* Should find more information from the whole content, including tables, text.

* For FF use 0.xx instead of xx.x, for example: 63.0 should return 0.63 !

for example, you should return:

```csv\\
Nickname,PCE\_max(\%),PCE\_ave(\%),Voc (V),Jsc (mA $cm^2$),FF\\
PBTTT-C14,2.34,2.34,0.53,9.37,0.48\\
```

Please use a step-by-step approach to analyze the content and ensure that all relevant information is accurately extracted. Only provide reasoning for how you identified each attribute and output the final csv format.

Format:

Reasoning: [Reasoning]

Answer: [Extracted csv]
\\
\\
\textbf{\small User Message:}

[ docoment.pdf ]

\end{prompt}

\subsubsection{Solubility Extraction}
\label{app:prompt-solubility-extraction}

Organic electrolytes, extensively used in battery technologies, comprise organic solvents, lithium salts, and additives. 
Understanding solubility in organic electrolytes is crucial as it impacts the efficiency of electrolytic processes, product selectivity, and equipment design. 
This task evaluates the LLM's capability in retrieving solubility-related tables. 
Papers typically select data from various aspects to describe the system, making it challenging to combine multiple tables for fuzzy matching. 
Therefore, we focus on examining the LLM's semantic understanding ability, enabling the model to select the most relevant and comprehensive table related to ``solubility'' from numerous alternatives and convert it into the specified format.

We prompt the model with the following:

\begin{prompt}{Prompt}

\textbf{\small System Message:}

You are an expert in the field of chemistry and specialize in the study of solubility.
Now you are required to extract tables related to solubility from the article.
The extracted information includes solute name, solvent name, temperature, pressure and solubility.
Since these properties are temperature-dependent and pressure-dependent, 
please place the properties at different temperatures or pressure on different rows.
The values of temperature and solubility should be output together with their unit.
Output the whole table in csv format and satisfy these requirements:

  (1) Do not truncate tables using "...". Always output the complete tables.
  
  (2) Keep all the superscripts in the form like "\textasciicircum 3", "\textasciicircum +" or "\textasciicircum a".

  (3) Do not use "NaN" to replace the blank cells, just leave it empty.
  
  (4) Use "x" to replace all "×", Use "()" to replace all " () "
  
  (5) Always add space before and after operators like " ± ".
  
As a example, the csv should be like:

```csv\\
solute\_name,solvent\_name,temperature,pressure,solubility\\
FLBDOB,PC,298.2 K,1 atm,0.275 ± 0.1 mol/L\\
```

Please use a step-by-step approach to analyze the content and ensure that all relevant information is accurately extracted. Only provide reasoning for how you identified each attribute and output the final csv format.

Format:

Reasoning: [Reasoning]

Answer: [Extracted csv]
\\
\\
\textbf{\small User Message:}

[ docoment.pdf ]

\end{prompt}

\subsubsection{Reactant QA}
\label{app:prompt-reactant-qa}

Organic and bio-catalyzed synthetic reactions are vital for the manufacture of drug-like molecules. 
Therefore, we designed a complex task to test the model's capability in extracting information from schematic diagrams and texts of chemical reactions. 
The model is required to understand the charts specified in the articles and select the correct answer from the provided multiple-choice descriptions.

\begin{prompt}{Prompt}
\textbf{\small System Message:}

You are an expert in the field of organic chemistry. Use chain of thought reasoning and provide reasoning before selecting the correct answer (e.g., a)xxx, or b)xxx).

Format:

Reasoning: [Reasoning]

Answer: [Answer]
\\
\\
\textbf{\small User Message:}

Which compound is in the reactants or reagents of the following reaction? 

a) c4ccc(B3OB(c1ccccc1)OB(c2ccccc2)O3)cc4\\
b) O=C(C(C)C(OC)=O)C1CC1\\
c) CC(=O)OP(=O)([O-])[O-].[NH4+].[NH4+]\\
d) COC(=O)/C(C)=C(/OS(=O)(=O)c1ccc(C)cc1)C1CC1 

The new reaction you should deal with is the second step in the first reaction in Section "2.2 Procedures".
\\
\\
\textbf{\small Expected Answer:}

b) O=C(C(C)C(OC)=O)C1CC1
\end{prompt}

\subsubsection{Reaction Mechanism QA}
\label{app:prompt-reaction-mechanism-qa}

Investigating electrolyte reactions helps improve the solid electrolyte interphase (SEI) layer, which directly affects battery performance and lifespan. 
Studies in this area lead to the development of advanced electrolytes that enhance a robust SEI, resulting in more efficient and durable batteries. 
We design a complex task to test the capability of extracting information from schematic diagrams of chemical reaction mechanisms. 
LLM is required to understand the specified reaction diagram and select the correct answer from the provided multiple choices.

We prompt the model with the following:
\begin{prompt}{Prompt}
\textbf{\small System Message:}

You are a highly intelligent organic electrolyte researcher who answers the following multiple choice question correctly.
Use chain of thought reasoning and provide reasoning before selecting the correct answer (e.g., a)xxx, or b)xxx).

Format:

Reasoning: [Reasoning]

Answer: [Answer]
\\
\\
\textbf{\small User Message:}

According to figure 1, which one of these synthetic routes for LTFOP is correct?

a) DTMSO + LiPF6 -> LTFOP + 2 CH3)3SiF

b) 2 DTMSO + LiPF6 -> LTFOP + 4 CH3)3SiF

c) HOOCCOOH + 2/3 CH3)3SiCl + 2/3 CH3)3SiNH)SiCH3) -> LTFOP + 2/3 NH4Cl

d) DTMSO + LiPCl6 -> LTFOP + 2 CH3)3SiCl
\\
\\
\textbf{\small Expected Answer:}

a) DTMSO + LiPF6 -> LTFOP + 2 CH3)3SiF

\end{prompt}

\subsection{Material}
\label{app:prompt-material}

\subsubsection{Material QA}
\label{app:prompt-material-qa}

The example of Material QA is similar to the MMLU-Pro-Biology task in Appendix \ref{app:prompt-mmlu-pro-biology}.








\subsubsection{Alloy Chart QA}
\label{app:prompt-alloy-chart-qa}

The processing steps and properties of alloy materials are often presented in charts, such as those comparing the performance of multiple alloys or illustrating how elongation changes with composition. 
Therefore, extracting information from these charts and integrating it with textual information is crucial. 
To further evaluate the retrieval capability of models regarding alloy chart information, we have designed multiple-choice questions involving alloy composition, processing techniques, and properties.

The example of Alloy Chart QA is similar to the Biology Chart QA task in Appendix \ref{app:prompt-biology-chart-qa}.

\subsubsection{Composition Extraction}
\label{app:prompt-composition-extraction}
Extracting alloy composition information from an article's text or tables and unifying it into a structured format helps researchers utilize historical data more effectively and provides valuable guidance for subsequent designs. 
This comprehensive task evaluates LLMs' ability to extract alloy compositions (including all element contents) from text and tables. 
Typically, alloy element content is found in two cases: (1) the element content is stored in a table, and (2) the element content is implicitly indicated by the alloy name, such as `Fe30Co20Ni50', which represents an atomic ratio of 30\% Fe, 20\% Co, and 50\% Ni. 
The objective of this task is to comprehensively extract this information and organize it into a digestible table.
The metric is to calculate the matching score between the standard answer table and the extraction result table. 
This task showcases the LLM's comprehension ability to integrate, extract, and structure multi-modal information \cite{kim2021prediction}.

An alloy composition table example is shown as following:
\begin{table}[t]
\centering
\tiny
\setlength{\tabcolsep}{0.9pt}
\begin{tabular}{cccccccc}
\toprule 
Alloy       & Composition & Composition    & Composition     & Composition    & Composition   & Composition  \\
\midrule
/               & C           & Cr             & Cu              & Fe             & Mn            & Mo           \\
\midrule
LeanDSS         & 0.014 \%    & 20.85 \%       & 0.09 \%         &  73.38 \%      & 1.49 \%       &   0.30 \%      \\
StandardDSS     & 0.012 \%    & 22.46 \%       & 0.17 \%         &  69.94 \%      & 1.81 \%       &   3.07 \%      \\
SuperDSS        & 0.013 \%    & 24.98 \%       & 0.20 \%         &  63.41 \%      & 0.48 \%       &   4.03 \%     \\
\bottomrule
\end{tabular}
\caption{Alloy composition example.}
\label{tab:alloy_sample}
\end{table}

\begin{prompt}{Prompt}
\textbf{\small System Message:}

You are an expert in the field of Alloy Materials.
Please give a complete list of alloy names and compositions of all alloys in this paper. 

If there is no alloy composition element ratio in the text, try to extract the element ratio from the alloy name from the perspective of alloy experts.

Output in csv format with multiindex (2 headers), The names in first header are 'AlloyName' and 'Composition' forcely. The names in second header are element names of alloy. 

Starting on the third row, list the alloy names and their corresponding element content. Based on the number of reference commas, the element name corresponds to the content.

Please write units not in header but in value like "50 wt.\%","30 at.\%". Output the data strictly in the CSV format shown below and exclude any other content. Example format:

```csv\\
AlloyName,Composition,Composition,Composition\\
nan,Fe,Co,Al\\
Fe70Co15Al3,70 wt.\%,15 wt.\%,3 wt.\%\\
Fe70Co18,70 wt.\%,18 wt.\%,nan\\
```

Please use a step-by-step approach to analyze the content and ensure that all relevant information is accurately extracted. Only provide reasoning for how you identified each attribute and output the final csv format.

Format:

Reasoning: [Reasoning]

Answer: [Extracted csv]
\\
\\
\textbf{\small User Message:}

[ docoment.pdf ]

\end{prompt}

\subsubsection{Temperature QA}
\label{app:prompt-temperature-qa}
The properties of an alloy are determined by its composition and the processes it undergoes, including processing and heat treatment. 
Therefore, extracting heat treatment values is critical. 
This task aims to determine the maximum temperature value for the heat treatment of the alloy. 
To ensure easy statistical analysis, questions are designed as multiple-choice.
Examples of process paragraphs \cite{villa2020physical}:

\begin{prompt}{Example Paragraph}
Cast NiMnGa samples, of Ni50Mn30Ga20 nominal composition, were prepared by 5 arc melting cycles of the pure elements (electrolytic Ni 99.97\%, electrolytic Mn 99.5\% and Ga 99.99\%) in stoichiometric ratio, in a non-consumable electrode furnace (Leybold LK6/45) (Leybold, Cologne, Germany). The as-cast ingot was ground to powder in a planetary ball mill (Fritsch Pulverisette 4) (FritschIdar-Oberstein, Germany) and the powder size was selected by means of sieves. Densified pellets were produced by die-pressing alloy powders with different average sizes (lower than 50 um or between 50 and 100 um) at 0.75 GPa at room temperature and sintered by thermal treatment at 925 \degree C for 24, 72, and 168 h in an Ar atmosphere, followed by slow cooling in the furnace. Sintered pellets had the following dimensions: approximately 3 mm in height and 13 mm in diameter. Table 1 provides a summary of the prepared sintered samples.
\end{prompt}

We prompt the model with the following:

\begin{prompt}{Prompt}
\textbf{\small System Message:}

You are an expert in the field of Alloy Materials. You are a highly intelligent alloy researcher who answers the following multiple choice question correctly.
Use chain of thought reasoning and provide reasoning before selecting the correct answer (e.g., a)xxx, or b)xxx).

Format:

Reasoning: [Reasoning]

Answer: [Answer]
\\
\\
\textbf{\small User Message:}

In the upper paper, what is the maximum temperature of the heat treatment process for all alloys?

a) 925 C

b) 650 C

c) 700 C

d) 800 C
\\
\\
\textbf{\small Expected Answer:}

a) 925 C
\end{prompt}

\subsubsection{Sample Differentiation}
\label{app:prompt-sample-differentiation}

Alloys with the same composition but treated by different processes are considered different samples because they exhibit different properties. 
Therefore, distinguishing between different samples and understanding the differences in their processes is essential. 
This multiple-choice question task is designed to comprehensively judge the number of different alloy samples proposed or studied by the authors. 
It assesses the LLMs' analysis and reasoning abilities regarding alloy distinctions from text.

The following example is process paragraphs where the sample are treated by different processes \cite{hernandez2017evaluation}:

\begin{prompt}{Example Paragraph}
An induction furnace was used to produce the Zn-21A1-2Cu alloy by melting proper amounts of Zn (99.99\%), Al (99.99\%), and Cu (99.96\%). The alloy was melted in a graphite crucible exposed to air and poured into cylindrical bars of 19 mm in diameter and 35 mm in length. After that, some bars were homogenized at 350 \degree C for 24h in the air. Cast and homogenized samples were subjected to an equal channel angular extrusion(ECAP) in a die with two cylindrical channels with a diameter of 15.8mm. The inner intersecting angle (y) was 90\ and the outer angle (y) was 36\degree. All samples were extruded by two and sixpasses with a ram velocity of 5 mm/min and by using B. route. The lubricant used was MoS, and it was applied to both channels on each pass.
\end{prompt}

We prompt the model with the following:

\begin{prompt}{Prompt}
\textbf{\small System Message:}

You are an expert in the field of Alloy Materials. Please answer the following multiple choice question correctly.
Use chain of thought reasoning and provide reasoning before selecting the correct answer (e.g., a)xxx, or b)xxx).

Format:

Reasoning: [Reasoning]

Answer: [Answer]
\\
\\
\textbf{\small User Message:}

Materials with the same components but processed through different techniques are considered as different alloys because they possess distinct properties. In the upper paper, please provide a count of all the alloys proposed and discussed by the authors?

a) 2

b) 0

c) 3

d) 1
\\
\\
\textbf{\small Expected Answer:}

d) 1
\end{prompt}

\subsubsection{Treatment Sequence}
\label{app:prompt-treatment-sequence}

Each alloy treatment process has a clear sequence requirement, so it is necessary to ensure that the extracted heat treatment process sequence is consistent with the experimental sequence.
For example, after solution treatment, a sample is further aged to ensure the release of internal stresses. 
This task aims to objectively analyze and evaluate the sequential relationship between two heat treatments and provide True/False answers. 
Additionally, if a specific heat treatment name does not exist in the paper, it should be considered False. 
This task assesses the LLM's comprehension ability to judge treatment order from the text.

\begin{prompt}{Prompt}
\textbf{\small System Message:}

You are an expert in the field of Alloy Materials. You are a specialist in the domain of heat treatment processes, such as homogenization, annealing, aging, solution treatment, quenching, and tempering, among others. Use chain of thought reasoning to analyze the question step by step. After your reasoning, answer the question with 'Yes' or 'No'.

Format:

Reasoning: [Reasoning]

Answer: [Yes/No]
\\
\\
\textbf{\small User Message:}

In the upper paper, is the processing heat treatment technique before the thermal treatment at 925 C called arc melting?
\\
\\
\textbf{\small Expected Answer:}

Yes
\end{prompt}

\subsection{Medicine}
\label{app:prompt-medicine}

\subsubsection{MMLU-Pro-Health}
\label{app:prompt-mmlu-pro-health}

The example of MMLU-Pro-Health is similar to the MMLU-Pro-Biology task in Appendix \ref{app:prompt-mmlu-pro-biology}.

\subsubsection{Affinity Extraction}
\label{app:prompt-affinity-extraction}

This task evaluates the LLM's ability to extract an affinity table containing molecules' tags, SMILES, and their affinities to different targets in bioassays. 
It tests several key capabilities of LLMs, including understanding complex and domain-specific language, as well as molecules and tables. 
Affinity data extraction requires not just surface-level text processing but also a deeper analysis to match different modalities.

An example output is shown in Table \ref{affinity}.

\begin{table*}[t]
\centering
\tiny
\begin{tabular}{>{\centering\arraybackslash}m{1.00cm} >{\centering\arraybackslash}p{0.5cm} >{\centering\arraybackslash}p{6.5cm} >{\centering\arraybackslash}p{2.5cm} >{\centering\arraybackslash}p{2.5cm}}
\hline
Compound & Name    & SMILES                & \multicolumn{2}{c}{Affinities}                                      \\
         &         &                       & Cytotoxicity in 2.2.15 Cells (IC50) & Anti-HBV Activity in 2.2.15 Cells (EC50) \\ \hline
         \addlinespace
1       & / & C1[C@H](O[C@H]([C@H]1F)\ N2C=NC3=C(N=CN=C32)N)CO & >200000 nM & >10000 nM \\
\addlinespace
2       & / & C1[C@H](O[C@H]([C@H]1F)\ N2C=CC(=NC2=O)N)CO & >200000 nM & 4000 nM \\ 
\addlinespace
3       & / & CC1=CN(C(=O)NC1=O)[C@H]2C\ [C@@H]([C@H](O2)CO)N=[N+]=[N-] & NA & NA \\
\hline
\end{tabular}
\caption{Example output of affinity data extraction task}
\label{affinity}
\end{table*}

We prompt the model with the following:

\begin{prompt}{Prompt}
\textbf{\small System Message:}

You are an expert in the field of pharmaceutical chemistry, and your task is to summarize the results of activity assays from an article in a tabular format. Please follow these steps to complete the task:

1. Determine if the article includes an activity assay. If it does, locate the section(s) presenting the assay results, which are usually in one or more tables.

2. Compile all the activity assay results into a single table. You may use multiple columns to represent different conditions or outcomes of various experiments.

3. Identify the names or codes used in the table, such as Example 1 or Compound A, and find the corresponding sections in the article that mention these substances. Extract the full name and SMILES notation of each substance.

4. Compile the names and SMILES notations of each substance in the table. Output in csv format with multiindex (Affinities, protein/cell line), write units not in header but in the value like "10.5 µM". Quote the value if it has comma! For example:

```csv\\
Compound,Name,SMILES,Affinities,Affinities,Affinities,Affinities\\
,,,5HT1A (IC50),5HT1D (IC50),5HT-UT (IC50),5HT1E (<affinity type>)\\
5a,Aspirin,CC(=O)Oc1ccccc1C(=O)O,2.0 nM,8.0 nM,12.6 nM, >1000 nM\\
```

5. If there are multiple tables, concat them. Don't give me reference or using "...", give me complete table!
Please use a step-by-step approach to analyze the content and ensure that all relevant information is accurately extracted. Only provide reasoning for how you identified each attribute and output the final csv format.

Format:

Reasoning: [Reasoning]

Answer: [Extracted csv]
\\
\\
\textbf{\small User Message:}

[ docoment.pdf ]

\end{prompt}

\subsubsection{Drug Chart QA}
\label{app:drug-chart-qa}

The analysis of drug properties, compositions, and processing techniques is critical for drug discovery and development. 
Often, this information is presented in charts, making it essential to extract and integrate such information with textual data. 
To further assess the retrieval capabilities of models in the context of drug chart information, we have designed multiple-choice questions focusing on drug composition, processing methods, and properties. 
The example of Drug Chart QA can be found in a similar format to the Biology Chart QA task in Appendix \ref{app:prompt-biology-chart-qa}.

\subsubsection{Tag to Molecule}
\label{app:prompt-tag2molecule}

This task evaluates the model's ability to find the correct SMILES given its tag in a document. 
Typically, a molecule is shown with an chart of its structure and a tag below it. 
The LLM should recognize both the structure and the tag and understand their connection.

\begin{prompt}{Prompt}
\textbf{\small System Message:}

You are an expert in the field of organic chemistry, can help user get SMILES formula from documents. Use chain of thought reasoning to analyze the document and extract the SMILES formula step by step. After reasoning, just give the SMILES formula as the final answer without further explanation.

Format:

Reasoning: [Reasoning]

Answer: [Extracted SMILES]
\\
\\
\textbf{\small User Message:}

What's the SMILES formula of molecule "Sumatriptan"?
\\
\\
\textbf{\small Expected Answer:}

"CNS(=O)(=O)CC1=CC2=C(C=C1)NC=C2CCN(C)C"
\end{prompt}

\subsubsection{Markush to Molecule}
\label{app:prompt-markush2molecule}

This task evaluates the model's ability to obtain the correct SMILES given a Markush formula (in CXSMILES pattern) and its substituents.

\begin{prompt}{Prompt}
\textbf{\small System Message:}

You are an expert in the field of chemistry, can help user insert substituents into CXSMILES-type markush formula to get SMILES formula (removing Hs). Use chain of thought reasoning to explain how you insert the substituents step by step, ensuring the correct SMILES is generated. After reasoning, just reply with the SMILES formula without further explanation.

Format:

Reasoning: [Reasoning]

Answer: [Generated SMILES]
\\
\\
\textbf{\small User Message:}

*C(*)CC(*)CC* |$A;;Pol_p;;;Q_e;;;M_p$|, A = H, Pol = NH2, Q = OH, M = [Li]
\\
\\
\textbf{\small Expected Answer:}

"NCCC(O)CC[Li]"
\end{prompt}

\subsubsection{Molecule in Document}
\label{app:prompt-molecule-in-doc}

This task evaluates the model's ability to determine whether a molecule (represented by SMILES) is mentioned in a document. 
The LLM should recognize all Markush formulas and their substituents, and then judge whether the required molecule is covered.

\begin{prompt}{Prompt}
\textbf{\small System Message:}

You are an expert in the field of chemistry. You are given a SMILES formula of a molecule, and should judge whether it is in the document. If the molecules are given by Markush formula (containing R group), You need to 1) analyze the skeletons of the provided molecule and the molecule in the literature or patent, and 2) if the compare the variable values of the molecular structure with the range of variable values given in the patent, to determine whether the molecule is covered by the literature or patent. Use chain of thought reasoning to analyze the question step by step. After your reasoning, answer the question with 'Yes' or 'No'.

Format:

Reasoning: [Reasoning]

Answer: [Yes/No]
\\
\\
\textbf{\small User Message:}

[ document.pdf ] 

Does the molecule "CC(CCCCCCCC1=CC(=C(C (=C1)OC)OC)OC)CCC(C2=CC=CS2)O" appear in the document?
\\
\\
\textbf{\small Expected Answer:}

Yes
\end{prompt}










\section{Baseline LLMs}
\label{app:baseline_model}

We briefly introduce the baseline LLMs and endpoints that we have tested on SciAssess.

\begin{itemize}[leftmargin=*]
    \item \textbf{OpenAI-o1}~\cite{o1}
    OpenAI's o1 model is designed to reason through complex tasks and solve harder problems in science, coding, and math.
    The model we tested is \texttt{OpenAI-o1-preview}.
    
    \item \textbf{GPT-4o}\footnote{\url{https://openai.com/index/hello-gpt-4o/}}: OpenAI's GPT-4o advances human-computer interaction by handling text, audio, image, and video inputs and outputs. It offers improved efficiency and cost compared to previous GPT models. The model we use is \texttt{gpt-4o}.

    \item \textbf{GPT-4} ~\cite{gpt4}: OpenAI's GPT-4 excels in text generation and comprehension, augmented with capabilities for image processing, code interpretation, and information retrieval. 
    These features make it adept at handling the complexities of scientific texts, positioning it as a versatile tool for scientific research. The model we use is \texttt{gpt-4-turbo}.
    
    \item \textbf{GPT-3.5}\footnote{\url{https://openai.com/blog/chatgpt}}: Preceding GPT-4, GPT-3.5 by OpenAI distinguishes itself with adept language processing skills, enabling effective engagement with complex texts. The model we use is \texttt{gpt-3.5-turbo-0125}.
    
    \item \textbf{Gemini-1.5-Pro}~\cite{gemini}: Google DeepMind's Gemini model family excels in multimodal comprehension, integrating text, code, image, and audio analysis. 

    \item \textbf{Claude 3 Opus}\footnote{\url{https://www.anthropic.com/news/claude-3-family}}: Claude 3 Opus model excels across major AI benchmarks, demonstrating near-human levels of comprehension and fluency in tasks like analysis, forecasting, and multilingual communication.

    \item \textbf{Moonshot-v1}\footnote{\url{https://platform.moonshot.cn/docs/intro}}: Moonshot-v1 is a text generation model proposed by Moonshot AI. We use \texttt{moonshot-v1-128k} in this study.

    \item \textbf{Doubao}\footnote{\url{https://www.volcengine.com/product/doubao}}: Doubao is a set of LLMs developed by ByteDance. The model we use is \texttt{Doubao-pro-128k}.

\end{itemize}

Apart from the closed-source LLMs, we also include some SOTA open-source LLMs:

\begin{itemize}[leftmargin=*]
    \item \textbf{Llama-3.1-70B}\footnote{\url{https://ai.meta.com/blog/meta-llama-3/}}: Llama 3-70B is a leading open-source LLMs released by Meta. 

     \item \textbf{Mixtral-8x22B}~\cite{mixtral}: Mixtral-8x22B-Instruct-v0.1 is the latest and largest mixture of experts large language model (LLM) from Mistral AI.
    
    \item \textbf{Qwen-2.5-72B}~\cite{qwen}: Qwen2 are series of LLMs developed by Alibaba. 
    The model we test is \texttt{Qwen2-72B-Instruct}.


\end{itemize}

\section{Performance without Cot}
\label{app:performance_wo_cot}

\begin{table*}[]
\centering
\tiny
\setlength{\tabcolsep}{2pt}
\begin{tabular}{>{
\centering\arraybackslash}m{1.25cm} | >{
\centering\arraybackslash}p{3cm} | >{
\centering\arraybackslash}p{1.00cm} >{
\centering\arraybackslash}p{0.85cm} >{
\centering\arraybackslash}p{0.85cm} >{
\centering\arraybackslash}p{0.85cm} >{
\centering\arraybackslash}p{0.85cm} >{
\centering\arraybackslash}p{0.85cm} >{
\centering\arraybackslash}p{0.85cm} >{
\centering\arraybackslash}p{0.85cm} | >{
\centering\arraybackslash}p{0.85cm} >{
\centering\arraybackslash}p{0.85cm} >{
\centering\arraybackslash}p{0.85cm}
}
\toprule
Domain & Task & o1-preview & GPT-4o & GPT-4 & GPT-3.5 & Moonshot & Claude3 & Doubao & Gemini & Llama3.1 & Qwen2.5 & Mixtral \\
\noalign{\vskip -1pt}\midrule\noalign{\vskip -1pt}
\multirow{6}{1.2cm}{\centering Biology}
& MMLU-Pro-Biology* & \textcolor{orange}{\textbf{0.901}} & 0.824 & 0.783 & 0.654 & 0.748 & 0.709 & 0.768 & 0.826 & 0.799 & \textcolor{teal}{\textbf{0.802}} & 0.709 \\
& Biology Chart QA & \textcolor{orange}{\textbf{0.653}} & 0.563 & 0.513 & 0.377 & 0.563 & 0.447 & 0.482 & \textcolor{orange}{\textbf{0.653}} & 0.503 & \textcolor{teal}{\textbf{0.533}} & 0.482 \\
& Chemical Entities Recognition* & \textcolor{orange}{\textbf{0.862}} & 0.855 & 0.845 & 0.614 & 0.803 & 0.826 & 0.786 & 0.799 & 0.824 & \textcolor{teal}{\textbf{0.845}} & 0.731 \\
& Compound Disease Recognition* & \textcolor{orange}{\textbf{0.745}} & 0.659 & 0.742 & 0.539 & 0.679 & 0.735 & 0.682 & 0.733 & 0.675 & \textcolor{teal}{\textbf{0.700}} & 0.605 \\
& Disease Entities Recognition* & \textcolor{orange}{\textbf{0.831}} & 0.809 & 0.828 & 0.697 & 0.715 & 0.797 & 0.789 & 0.822 & \textcolor{teal}{\textbf{0.814}} & 0.700 & 0.783 \\
& Gene Disease Function* & 0.687 & 0.488 & 0.717 & 0.523 & 0.637 & 0.558 & 0.649 & \textcolor{orange}{\textbf{0.792}} & \textcolor{teal}{\textbf{0.655}} & 0.495 & 0.484 \\
\noalign{\vskip -1pt}\midrule\noalign{\vskip -1pt}
\multirow{9}{1.2cm}{\centering Chemistry} 
& MMLU-Pro-Chemistry* & \textcolor{orange}{\textbf{0.868}} & 0.339 & 0.334 & 0.228 & 0.260 & 0.226 & 0.342 & 0.513 & 0.347 & \textcolor{teal}{\textbf{0.420}} & 0.314 \\
& Electrolyte Table QA & \textcolor{orange}{\textbf{0.925}} & 0.590 & 0.380 & 0.170 & 0.735 & 0.315 & 0.640 & 0.845 & 0.490 & \textcolor{teal}{\textbf{0.668}} & 0.337 \\
& OLED Property Extraction & 0.394 & \textcolor{orange}{\textbf{0.459}} & 0.390 & 0.107 & 0.165 & 0.130 & 0.327 & 0.365 & 0.103 & \textcolor{teal}{\textbf{0.371}} & 0.237 \\
& Polymer Chart QA & \textcolor{orange}{\textbf{1.000}} & 0.867 & 0.600 & 0.067 & 0.733 & 0.133 & 0.733 & 0.867 & \textcolor{teal}{\textbf{0.867}} & \textcolor{teal}{\textbf{0.867}} & 0.800 \\
& Polymer Composition QA & \textcolor{orange}{\textbf{0.986}} & 0.756 & 0.708 & 0.316 & 0.967 & 0.608 & 0.823 & 0.952 & 0.689 & \textcolor{teal}{\textbf{0.894}} & 0.524 \\
& Polymer Property Extraction & 0.606 & \textcolor{orange}{\textbf{0.785}} & 0.782 & 0.435 & 0.705 & 0.489 & 0.524 & 0.701 & 0.590 & \textcolor{teal}{\textbf{0.689}} & 0.559 \\
& Solubility Extraction & 0.427 & 0.508 & \textcolor{orange}{\textbf{0.516}} & 0.326 & 0.476 & 0.396 & 0.407 & 0.454 & 0.443 & \textcolor{teal}{\textbf{0.448}} & 0.303 \\
& Reactant QA & \textcolor{orange}{\textbf{0.559}} & 0.374 & 0.359 & 0.251 & 0.256 & 0.226 & 0.241 & 0.338 & 0.277 & \textcolor{teal}{\textbf{0.472}} & 0.205 \\
& Reaction Mechanism QA & 0.682 & 0.545 & 0.500 & 0.091 & 0.591 & 0.409 & 0.409 & \textcolor{orange}{\textbf{0.773}} & \textcolor{teal}{\textbf{0.682}} & 0.455 & 0.409 \\
\noalign{\vskip -1pt}\midrule\noalign{\vskip -1pt}
\multirow{6}{1.2cm}{\centering Material} 
& Material QA & \textcolor{orange}{\textbf{0.821}} & 0.757 & 0.707 & 0.521 & 0.586 & 0.544 & 0.665 & 0.715 & 0.684 & \textcolor{teal}{\textbf{0.738}} & 0.654 \\
& Alloy Chart QA & 0.533 & 0.467 & 0.533 & 0.333 & 0.333 & 0.467 & \textcolor{orange}{\textbf{0.733}} & 0.667 & 0.467 & \textcolor{teal}{\textbf{0.667}} & 0.600 \\
& Composition Extraction & \textcolor{orange}{\textbf{0.488}} & \textcolor{orange}{\textbf{0.488}} & 0.441 & 0.100 & 0.351 & 0.360 & 0.336 & 0.423 & 0.424 & \textcolor{teal}{\textbf{0.433}} & 0.198 \\
& Temperature QA & 0.836 & 0.609 & 0.536 & 0.261 & 0.836 & 0.295 & 0.425 & \textcolor{orange}{\textbf{0.879}} & \textcolor{teal}{\textbf{0.594}} & 0.507 & 0.353 \\
& Sample Differentiation & 0.392 & 0.245 & 0.333 & 0.089 & \textcolor{orange}{\textbf{0.662}} & 0.274 & 0.207 & 0.641 & 0.211 & 0.194 & \textcolor{teal}{\textbf{0.300}} \\
& Treatment Sequence & 0.624 & 0.599 & 0.421 & 0.431 & \textcolor{orange}{\textbf{0.683}} & 0.470 & 0.564 & 0.634 & 0.604 & \textcolor{teal}{\textbf{0.629}} & 0.411 \\
\noalign{\vskip -1pt}\midrule\noalign{\vskip -1pt}
\multirow{6}{1.2cm}{\centering Medicine} 
& MMLU-Pro-Health* & \textcolor{orange}{\textbf{0.784}} & 0.759 & 0.681 & 0.498 & 0.603 & 0.549 & 0.634 & 0.647 & \textcolor{teal}{\textbf{0.686}} & 0.660 & 0.532 \\
& Affinity Extraction & 0.068 & \textcolor{orange}{\textbf{0.104}} & 0.074 & 0.051 & 0.041 & 0.027 & 0.065 & 0.087 & \textcolor{teal}{\textbf{0.086}} & 0.058 & 0.052 \\
& Drug Chart QA & \textcolor{orange}{\textbf{0.600}} & 0.467 & 0.333 & 0.400 & 0.333 & 0.333 & 0.333 & 0.400 & \textcolor{teal}{\textbf{0.400}} & \textcolor{teal}{\textbf{0.400}} & 0.267 \\
& Tag2Mol & 0.127 & 0.073 & 0.036 & 0.000 & 0.127 & 0.008 & 0.123 & \textcolor{orange}{\textbf{0.216}} & \textcolor{teal}{\textbf{0.097}} & 0.002 & 0.000 \\
& Markush2Mol & 0.662 & 0.645 & \textcolor{orange}{\textbf{0.675}} & 0.488 & 0.664 & 0.519 & 0.583 & 0.671 & 0.376 & \textcolor{teal}{\textbf{0.475}} & 0.400 \\
& Mol In Document & \textcolor{orange}{\textbf{0.840}} & 0.480 & 0.580 & 0.380 & 0.460 & 0.460 & 0.600 & 0.700 & 0.480 & \textcolor{teal}{\textbf{0.540}} & 0.460 \\
\bottomrule
\end{tabular}
\caption{Performance comparison of LLMs across various scientific domains. 
\textcolor{orange}{\textbf{Orange}} and \textcolor{teal}{\textbf{green}} indicate the best in closed and open source LLMs, respectively.
* indicates 3-shot.
The prompts simply require the model to return the final answer without chain-of-thought.}
\label{tab:all_result_without_cot}
\end{table*}